%% file: root.tex
\documentclass[letterpaper, 10 pt, conference]{ieeeconf}  

\IEEEoverridecommandlockouts                              

\usepackage{graphics} 
\usepackage{amsmath} 
\usepackage{tabularx}
\usepackage{booktabs}
\usepackage{algorithm}
\usepackage{algorithmic}
\usepackage{subcaption}
\usepackage[dvipsnames]{xcolor}
\usepackage{graphicx}
\graphicspath{ {./imgs/} }
\usepackage{hyperref}

\title{Leveraging Distributional Bias For Reactive Collision Avoidance under Uncertainty: A Kernel Embedding Approach}
\author{Anish Gupta$^1$, Arun Kumar Singh$^2$, and K. Madhava Krishna$^1$ 
\thanks{1. Robotics Research Center, IIIT Hyderabad, India 2. University of Tartu, Estonia. This work was supported in part by the European Social Fund via ICT programme measure, Estonian Center of Excellence in IT (EXCITE) funded by the European Regional Development Fund, grant PSG753 from Estonian Research Council, and Artificial Intelligence \& Robotics Estonia (AIRE), the Estonian candidate for European Digital Innovation Hub, funded by the Ministry of Economic Affairs and Communications in Estonia.}
}

\newcommand{\norm}[1]{\left\lVert#1\right\rVert}

\begin{document}
\maketitle
\thispagestyle{empty}
\pagestyle{empty}

\input{chapters/abstract}
\input{chapters/introduction}

\input{chapters/related-work}

\input{chapters/methodology}
\input{chapters/results}

\input{chapters/conclusion}
\bibliographystyle{IEEEtran}
\bibliography{references} 
\end{document}

%% file: chapters/abstract.tex
\begin{abstract}
Many commodity sensors that measure the robot and dynamic obstacle's state have non-Gaussian noise characteristics. Yet, many current approaches treat the underlying uncertainty in motion and perception as Gaussian, primarily to ensure computational tractability. On the other hand, existing planners working with non-Gaussian uncertainty do not shed light on leveraging distributional characteristics of motion and perception noise, such as bias for efficient collision avoidance. 

This paper fills this gap by interpreting reactive collision avoidance as a distribution matching problem between the collision constraint violations and Dirac Delta distribution. To ensure fast reactivity in the planner, we embed each distribution in Reproducing Kernel Hilbert Space and reformulate the distribution matching as minimizing the Maximum Mean Discrepancy (MMD) between the two distributions. We show that evaluating the MMD for a given control input boils down to just matrix-matrix products. We leverage this insight to develop a simple control sampling approach for reactive collision avoidance with dynamic and uncertain obstacles. 

We advance the state-of-the-art in two respects. First, we conduct an extensive empirical study to show that our planner can infer distributional bias from sample-level information.
Consequently, it uses this insight to guide the robot to good homotopy. We also highlight how a Gaussian approximation of the underlying uncertainty can lose the bias estimate and guide the robot to unfavorable states with a high collision probability. Second, we show tangible comparative advantages of the proposed distribution matching approach for collision avoidance with previous non-parametric and Gaussian approximated methods of reactive collision avoidance.




\end{abstract}

%% file: chapters/introduction.tex
\section{Introduction} \label{Intro}
Collision avoidance under uncertainty has been well studied in existing literature \cite{mora-cco, manocha, PVO}. Most of them assume Gaussian perturbation in estimating robot and obstacles' state and motion commands executed by the robot for collision avoidance. The primary motivation for Gaussian approximation is that in many cases, this leads to efficient convex structures in the problem \cite{alonso_cco}. However, when the underlying uncertainty is very far from Gaussian, such approximations adversely affect planning efficiency by providing a conservative estimate of the feasible space. 

\begin{figure}
        \includegraphics[scale = 0.25]{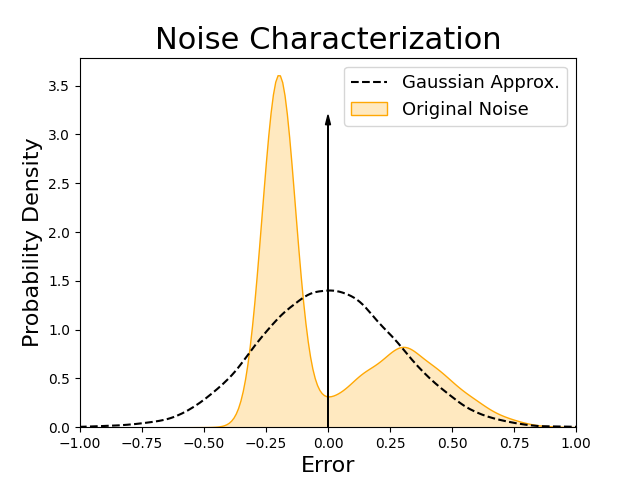}
        \includegraphics[scale = 0.25]{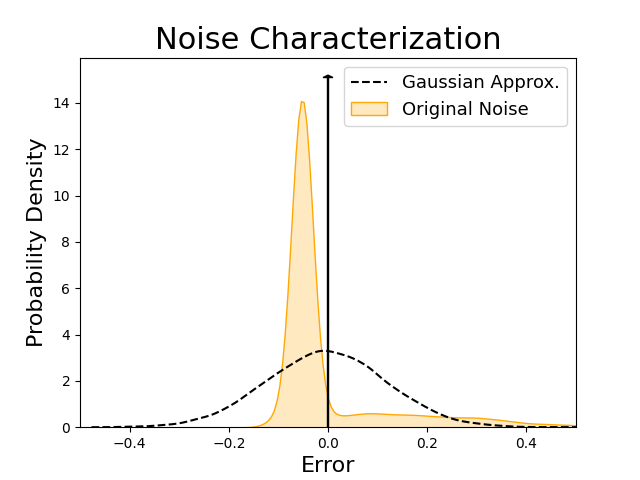}
         \caption{\small Examples of Non-Gaussian distribution and their Gaussian approximations. The majority of the mass of the true distribution is shifted with respect to the mean of the Gaussian approximation. We refer to it as the distribution bias.}
    \label{noise}
\end{figure}

\begin{figure}
        \includegraphics[scale = 0.15]{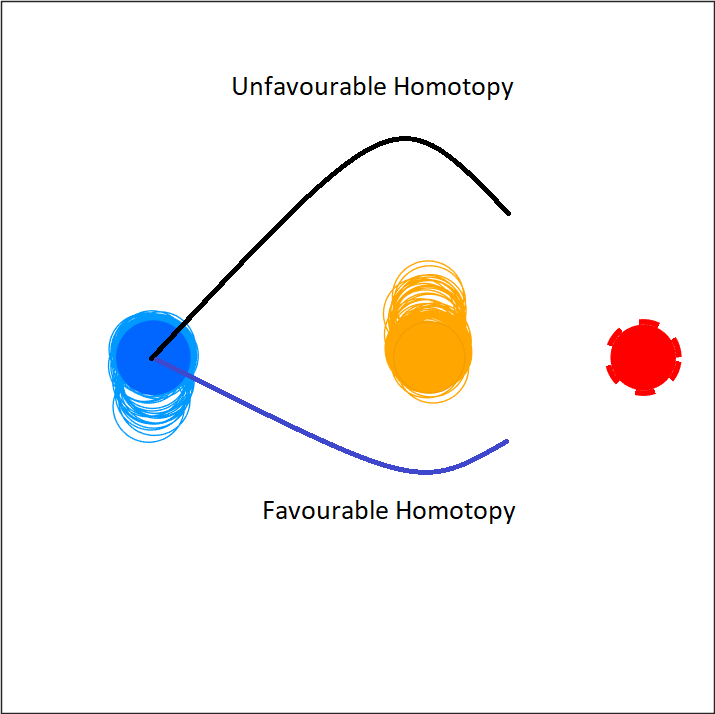}
        \hspace{0.5 cm}
        \includegraphics[scale = 0.15]{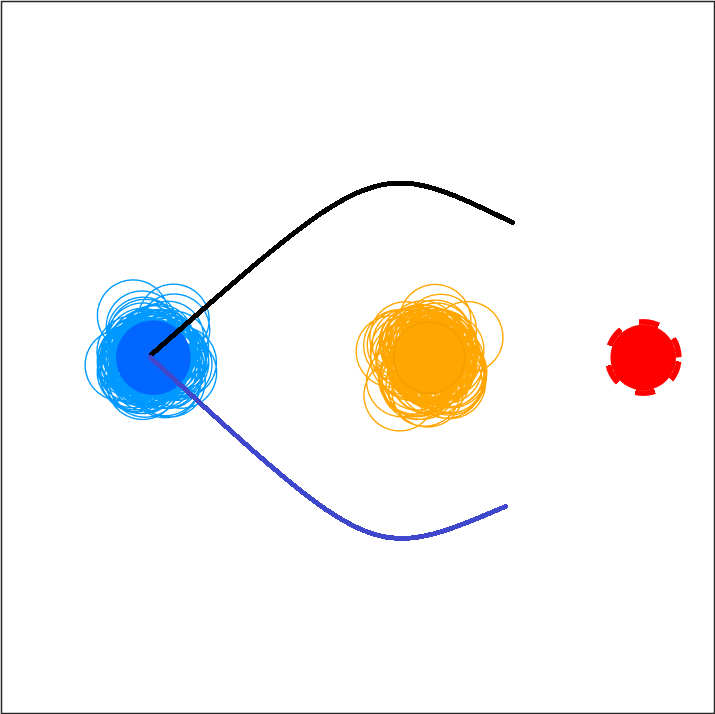}
        \caption{\small{Left figure shows collision avoidance under non-Gaussian motion  and perception noise. The goal position is shown in red. The robot (blue) can choose to avoid the obstacle (orange) from either left or right. However, due to the presence of bias in the motion and perception noise, one of the homotopies shown in blue becomes more favorable. Our objective in this paper is to develop reactive planners than can guide the robots towards favorable homotopies. The right figure presents the situation under Gaussian approximation of the noise. In this case, either homotopy erroneously appear equally good (or worse). Thus, it is quite likely that Gaussian approximation will lead the robot unfavorable positions with high collision probability }  }
    \label{teaser}
    \vspace{-0.8cm}
\end{figure}

Recent works like \cite{leonard, anirudha} are capable of planning and control under non-Gaussian noise motion and perception noise. Some recent approaches like \cite{bharath_ral, bharath_tcst, manocha} specifically deal with reactive collision avoidance in a similar vein to our current work. These methods highlight the reduction in collision probability and control effort achieved by adopting a more sophisticated vies of the underlying uncertainty. However, they do not provide a fine-grained analysis of how distributional characteristics like bias affect collision avoidance and how we can leverage it to reduce collision probabilities. The bias we refer to can be described through figure \ref{noise} wherein a bi-modal distribution obtained from a commodity GPS is not mean-centered. In other words, the presence of the second mode provides for an unequal spread on either side of the mean, unlike its Gaussian approximation. As shown in Fig.\ref{teaser}, the unequal spread of distribution mass naturally creates a notion of favorable and unfavorable homotopies. The former corresponds to the side where there would be less overlap between robot and obstacles uncertainty. To the best of our knowledge, no existing approaches can ensure that the robot chooses the favorable homotopies with high probability while avoiding obstacles. In this paper, we fill this current knowledge gap by analyzing in diverse ways why a particular control action is chosen for a given obstacle configuration and how they depend on the nature of the underlying uncertainty and any approximation we make on it. Our work uses the template of chance-constrained optimization (CCO) and its core novelties can be summarized as follows.

\noindent \textbf{Algorithmic Contribution:} Our main hypothesis in this work is that bias in the motion and perception noise can actually be leveraged for efficient planning. However, this requires reactive planners capable of operating under non-parametric uncertainty. To this end, we follow our prior work \cite{bharath_ral, bharath_tcst, cco_voxel} which interprets CCO as a distribution matching problem. Specifically, we reformulate CCO as the problem of finding the appropriate control inputs that minimizes the deviation between the violation of velocity obstacle (VO) constraints and Dirac-Delta distributions. We show how we can leverage distribution embedding in Reproducing Kernel Hilbert Space (RKHS) to formulate distribution matching cost as Maximum Mean Discrepancy (MMD) measure. Moreover, we use the so-called kernel-trick to reduce MMD evaluation for a given control input to computing a few matrix-matrix products. This allows us to adopt a simple control sampling approach for real-time reactive navigation. 

\noindent \textbf{Empirical Contribution:} For the first time, we show the importance of retaining the true non-parametric nature of the distribution while computing motion plans. In particular, we show that given an appropriate planner (e.g the current proposed one), the distributional bias can be leveraged for drastically improving collision probability and control effort by guiding robots towards favorable homotopies. In contrast, when we approximate the uncertainty as Gaussian, the planner's choice of homotopy is completely random leading to higher collision probabilities. 

\noindent  \textbf{Benchmarking Contribution:} We compare our planner to a diverse set of baselines and show significant improvement in collision probabilities and control costs. Our first baseline is \cite{prvo} that uses Gaussian approximation of motion and perception uncertainty to obtain tractable reformulation of the chance constraints. Our second baseline follows the same distribution matching interpretation of CCO as ours but fits Gaussian Mixture Model to the uncertainty and uses Kullback Liebler Divergence (KLD) to construct distribution matching cost. Our final baseline is infact, an ablation where we use our MMD based approach but with Gaussian approximation of the uncertainty.

%% file: chapters/related-work.tex
\section{Related Work} \label{RW}

Chance constrained optimization has emerged as the popular paradigm and framework for collision avoidance under uncertainty \cite{mora-cco, manocha}. While there are many variants of this problem typical formulations recast the original intractable chance constraints into surrogate forms. In special cases such as in \cite{IROS-15} closed form solutions are possible. Most such formulations model the original distribution to be a Gaussian and resort to linearization \cite{mora-cco} or manage closed form solutions when the collision avoidance constraints can be posed as convex or affine constraints \cite{ecc14,ecc15}. Methods such as \cite{ecc16,ecc17,ecc18} have formulated surrogates that give tight approximations to these chance constraints defined over non-linear inequalities(could be non-convex also), whereas in \cite{PVO} a Bayesian Decomposition framework is proposed in a multi-robotic setting. However, all these algorithms require a fundamental assumption on the nature of uncertainty of the random variables (state and actuation of the robot) involved . In general closed-form, surrogates can only be derived if the random variables belong to a Gaussian distribution.

Recently there has been a growing trend towards non-parametric chance constraints \cite{bharath_ral, bharath_tcst, manocha} acknowledging that most sensor noise are more often non-parametric \cite{bharath_ral, anirudha}. These methods typically pose chance constraint optimization as one of distribution matching either through the popular KL distance or by computing distance between distributions in their Hilbert Space embedding. While showcasing efficacious outcomes in terms of various metrics and rubrics these methods have not indulged in an analysis that pinpoints how and why non parametric modeling actually is beneficial and why Gaussian approximations can provide for inaccurate control actions and outcomes.

In this paper we try to close this gap by providing detailed empirical analysis as to how the inherent bias prevalent in non-parametric distribution can pose a challenge when approximated by parametric Gaussian noise. We also contribute over prior works \cite{bharath_ral}, \cite{ecc20} by circumventing the need for estimating the desired distribution in the distributon matching interpretation of CCO. To this end, we also extend \cite{cco_voxel} to reactive navigation in dynamic environments.

%% file: chapters/methodology.tex
\section{Preliminaries and Problem Formulation} \label{pf}
\noindent \textbf{Symbols and Notations:} We represent scalars as normal case small font letters and use the bold font variant for vectors. We use bold-font upper case letters to denote matrices. We use subscript $t$ to denote the time-stamp of a variable. The notation $\Vert \cdot \Vert$ denotes the Euclidean norm of vector/matrices. We use $\overline{c}$ to denote some nominal/noise-free value of a random variable $c$. The symbol $\Pr(\cdot)$ denotes the probability of an event, while $p(.)$ represents the probability distribution function. Some of the commonly used symbols and notations are summarized in the table \ref{tab:notations} while some are also defined in their first place of use. 

\begin{table}
\caption{}
\label{tab:notations}
\begin{tabular}{@{}ll@{}}
\toprule
\multicolumn{1}{c}{\textbf{Symbol}} & \multicolumn{1}{c}{\textbf{Description}} \\ \midrule
$(\textbf{x}_t, \textbf{v}_t)$ & Position and linear velocity of the robot at time $t$ \\ \midrule
$(\theta_t, \omega_t)$ & Heading and angular velocity of the robot at time $t$ \\ \midrule
$(\textbf{x}_{o, t}, \textbf{v}_{o, t})$ & Position and velocity of the obstacle at time $t$ \\ \midrule
$\textbf{u}_t$ & Control input to the robot at time $t$ \\ \midrule
$f(\cdot) \leq 0$ & VO constraints for $j^{th}$ obstacle \\ \midrule
$p_{f}$ & Distribution of $f(.)$ under perception and ego uncertainty \\ \midrule
$\eta$ & Probability of collision avoidance \\ \bottomrule
\end{tabular}
\end{table}
\subsection{Motion and Perception Model}
\label{sec:robot-state-estimation}
\noindent We assume that the robot has the following discrete time stochastic motion model, wherein $\Delta t$ represents the time duration between two consecutive steps.

\begin{align}
    \textbf{x}_{t+1} = \textbf{x}_t+\textbf{v}_t\Delta t, \theta_{t+1} = \theta_t+\omega_t\Delta t,\\
    \textbf{v}_t = \begin{bmatrix}
            v_t \cos(\theta_t + \omega_t \Delta t)\\
            v_t \sin(\theta_t + \omega_t \Delta t)
        \end{bmatrix},\\
          \begin{bmatrix}
            v_t \\
            \omega_t
        \end{bmatrix} = \overbrace{\begin{bmatrix}
            \overline{v}_t \\
            \overline{\omega}_t  
        \end{bmatrix}}^{\textbf{u}_t}+\boldsymbol{\epsilon} \label{control_corrupted}
\end{align}
In the stochastic setting, $\textbf{x}_t, \theta_t, v_t, \omega_t$ are all random variables with unknown probability distribution. To simplify the technical exposition, we assume that these variables have a nominal noise-free value corrupted by an additive disturbance. For example, as shown in \eqref{control_corrupted}, the control consists of deterministic command of linear ($\overline{v}_t$) and angular ($\overline{\omega}_t$) velocity corrupted by $\boldsymbol{\epsilon}$. Although, the probability distribution of the disturbance is not known, we assume to have access to the samples drawn from it. We also assume that a Particle filter like set-up is in place that bounds the uncertainty in position at each time step.

We represent obstacles' motions through the following piece-wise straight line trajectory. Similar to robot motion model, we treat $\textbf{x}_{oj, t}, \textbf{v}_{oj, t} $ as random variables and we have access to only small number of sample realizations of these random variables.

\begin{align}
    \textbf{x}_{o, t+1} = \textbf{x}_{o, t}+\textbf{v}_{o, t}\Delta t
\end{align}

\subsection{Velocity Obstacle Constraints}
\label{sec:collision-avoidance-chance-constraints}
\label{sec:vo}

\noindent In deterministic noise-less setting, reactive collision avoidance between disk shaped robots and obstacles is often formulated in terms of velocity obstacle (VO) \cite{vo} constraints defined in the following manner:
    \begin{subequations}
    \label{eq:vo}
    \begin{equation}
        f(\cdot) \leq 0 : \frac{(\textbf{r}^T \textbf{v})^2}{\norm{\textbf{v}}^2} - \norm{\textbf{r}}^2 + R^2 \leq 0, \forall j
    \end{equation}
    \begin{equation}
        \textbf{r} = {\textbf{x}_t} - {\textbf{x}_{o, t}},\ \ 
        \textbf{v} = \textbf{v}_t - \textbf{v}_{o, t}
    \end{equation}
\end{subequations}

\noindent where, $R$ represents the combined radii of the robot and the obstacle. For the ease of exposition in the latter sections, we formulate the VO constraints for a single obstacle. Extension to multiple obstacles is trivial.

In the stochastic setting where the robot and obstacle's position and velocities are all random variables, the function $f(.)$ actually characterizes a distribution. Thus, we need to suitably modify our use of VO constraints. This is discussed next.




\subsection{Reactive Navigation Through CCO}
\label{sec:problem-formulation}
\noindent We formulate one-step reactive navigation in uncertain environment as the following CCO:

\begin{subequations}
    \label{eq:mpc}
    \begin{equation}
        \label{eq:mpc-cost}
        \min_{\textbf{u}_t} w_1\Vert \overline{\textbf{v}}_t-\textbf{v}_d\Vert_2^2 + w_2 \norm{\overline{\textbf{u}}_t}^2
    \end{equation}
    \begin{equation}
        \label{eq:mpc-chance-constraints}
    \Pr(f(\textbf{x}_t, \theta_t ,  \textbf{u}_t, \textbf{x}_{o, t}, \textbf{v}_{o, t}) \leq 0) \geq \eta,\ \forall j,\  \textbf{u}_t \in \mathcal{C}
    \end{equation}
\end{subequations}

\begin{align}
    \overline{\textbf{v}}_t = \begin{bmatrix}
            \overline{v}_t \cos(\overline{\theta}_t + \overline{\omega}_t \Delta t)\\
            \overline{v}_t \sin(\overline{\theta}_t + \overline{\omega}_t \Delta t)
        \end{bmatrix},
\end{align}

\noindent The first term in the cost function \eqref{eq:mpc-cost} ensures that the nominal velocity is aligned with some desired velocity vector $\textbf{v}_d$, which is often constructed in a way to induce movement towards the goal \cite{manocha-rvo}. We also have a regularizer on the control input in the cost function \eqref{eq:mpc-cost}. The weights $w_1$ and $w_2$ are user defined and used to balance the trade-off between each cost term. The $\mathcal{C}$ represents the set of feasible control inputs and we assume that it is convex formed by affine constraints on $\overline{v}_t, \overline{\omega}_t$. The set of inequalities \eqref{eq:mpc-chance-constraints} represent the so-called chance constraints \cite{bharath_tcst} and ensures that the probability of satisfaction of VO constraints is greater than or equal to some threshold $\eta$.

The main computational challenge in solving (\ref{eq:mpc-cost})-\eqref{eq:mpc-chance-constraints} stems from the chance constraints. Thus, existing works heavily focus on replacing \eqref{eq:mpc-chance-constraints} with more computationally tractable options. We discuss one of them in the Section \ref{Results}. However, most of the existing reformulations assume that the underlying uncertainty is Gaussian \cite{alonso_cco}. Since our aim in this paper is to analyze the effect of Gaussian approximation, we next present our reactive planner that can work with arbitrary uncertainty distribution.

\section{Methods}


\subsection{Reformulation as a Distribution Matching Problem}
\noindent At an intuitive level CCO \eqref{eq:mpc-cost}-\eqref{eq:mpc-chance-constraints} has the following interpretation \cite{bharath_tcst}. We seek to compute a nominal control $\overline{\textbf{u}}_t$ that modifies the shape of the distribution of $f(.)$ in  a way that most of its mass lies on the left of the line $f_j(.)=0$. An alternate interpretation can be derived by defining a function $h$ in the following manner.

\small
\begin{align}
    h(\textbf{x}_t, \theta_t ,  \textbf{u}_t, \textbf{x}_{o, t}, \textbf{v}_{o, t}) = \max(0, f(\textbf{x}_t, \theta_t ,  \textbf{u}_t, \textbf{x}_{o, t}, \textbf{v}_{o, t}))
    \label{const_viol_func}
\end{align}
\normalsize

As clear, $h(.)$ measures constraint violation. It is zero if the VO constraints are satisfied and equal to $f(.)$ otherwise. In the stochastic setting where $\textbf{x}_t, \textbf{u}_t, \textbf{x}_{oj, t}, \textbf{v}_{oj, t}$ are random variables, $h(.)$ defines the distribution of constraint violations. 

With respect to \eqref{const_viol_func}, we can interpret CCO as the problem of finding an appropriate control input $\overline{\textbf{u}}_t$ such that the distribution of $h(.)$ becomes similar to that of a Dirac-Delta. Using this interpretation, we can reformulate \eqref{eq:mpc-cost}-\eqref{eq:mpc-chance-constraints} in the following manner:

\begin{align}
    \label{final_cost}
    \min_{\overline{\textbf{u}}_t} l_{dist}(p_{h}, p_{\delta})+ w_1\Vert \overline{\textbf{v}}_t-\textbf{v}_d\Vert_2^2 + w_2\norm{\overline{\textbf{u}}_t}^2\\
    \textbf{u}_t \in \mathcal{C}, \label{feas_final}
\end{align}
\noindent where $p_{h}, p_{\delta}$ represents the probability distribution of $h(.)$ and Dirac-Delta respectively.  The function $l_{dist}$ measures the similarity between $p_{h}, p_{\delta}$ and it decreases as the distribution becomes similar. One possible option for $l_{dist}$ is the KL divergence. However, it cannot operate at purely sample level and requires the parametric form of the distributions to be known. Thus, we define $l_{dist}$ as MMD between $p_h$ and $p_{\delta}$ defined in the following manner.

\begin{align}
   l_{dist}(\textbf{u}_t) =  \overbrace{\Vert \mu_{p_{h}}(\textbf{u}_t) -\mu_{p_{\delta}}\Vert_2^2}^{MMD},
    \label{l_dist}
\end{align}

\noindent where, $\mu_{p_{h}}$ and $\mu_{p_{\delta}}$ represent the RKHS embedding of $p_{h}$ and $p_{\delta}$ respectively. 

We solve \eqref{final_cost}-\ref{feas_final} through a simple control sampling approach. We draw several samples of $\textbf{u}_t$ from a  uniform distribution and then evaluate the cost \eqref{final_cost} on them. Subsequently, we choose the sample corresponding to the lowest cost. Our control sampling relies on efficient evaluation of MMD term to retain online performance. Thus, in the next section, we show how MMD evaluation for a given $\textbf{u}_t$ can be reduced to computing matrix-matrix products.

\subsection{Matrix Representation for MMD}
\noindent The algebraic expression for $\mu_{p_{h}}$  can be derived in the following manner

\begin{align}
    \mu_{p_{h}} = \sum_{i=0}^{i=N} \sum_{j=0}^{j=N} \alpha_i \beta_j k(\textbf{h}_{ij}, .)
    \label{kernel_mu_h}
\end{align}
\noindent where 

\begin{align}
    h_{ij} = h(\textbf{x}_t^i, \theta_t^i ,  \textbf{u}_t, \textbf{x}_{o, t}^j, \textbf{v}_{o, t}^j)
\end{align}

\noindent  and $\alpha_i, \beta_j$ are constants. Typically, if we draw I.I.D samples, then we have $\alpha_i=\beta_j = \frac{1}{n}$. However, as shown in \cite{bharath_tcst}, \cite{bharath_ral}, these constants can be chosen in a clever way to re-weight the importance of each samples leading to sample efficiency. The function $k(., .)$ is the so-called kernel operator, which in our implementation as Radial Basis Function. That is, $k(\textbf{c}_1, \textbf{c}_2) = exp(-\gamma \Vert \textbf{c}_1-\textbf{c}_2\Vert_2^2 )$ for some arbitrary vectors $\textbf{c}_1, \textbf{c}_2$. 

As clear, $\mu_{p_{h}}$ is formed by first drawing $n$ samples each of robot position/heading $(\textbf{x}_t^i, \theta_t^i)$ and obstacle position/velocity ($(\textbf{x}_{o, t}^j, \textbf{v}_{o, t}^j)$) distribution and then evaluating $h(.)$ over all the possible sample pairs. The function $k(.)$ represents the feature map associated with the RBF kernel. We can represent \eqref{kernel_mu_h} in the following more compact form, wherein $a_p$ denotes the $p^{th}$ element of the vector $\textbf{a}$

\begin{align}
    \label{cone_dist}
    \mu_{p_{h}} = \sum_{p=1}^{p=N^2}a_p k(\textbf{h}_p,.),  \textbf{h}_p = \begin{bmatrix}
      h_{11}\\
      h_{12}\\
      \vdots\\
      h_{nn}
    \end{bmatrix}, \textbf{a} = \begin{bmatrix}
      \alpha_1\beta_1\\
      \alpha_1\beta_2\\
      \vdots\\
      \alpha_n\beta_n
    \end{bmatrix}
\end{align}

Following a similar approach, we can define $\mu_{p_{\delta}}$ as 

\begin{align}
    \mu_{p_{\delta}} = \sum_{q=1}^{q=N^2}b_q k(0,.)
    \label{mu_delta}
\end{align}

\noindent for some constant vector $\textbf{b}$. Note that \eqref{mu_delta} exploits the fact that the samples from a Dirac-Delta distribution are all zeros.

With respect to the above definition, we can expand \eqref{l_dist} as

\begin{subequations}
    \begin{align}
        \norm{\mu_{p_{h}}-\mu_{P_{\delta}}}^2_2=  \textbf{M}_{cc} - 2\textbf{M}_{c0} + \textbf{M}_{00}\\
        \text{where, \hspace{0.4 cm}} \textbf{M}_{cc} = \langle \mu_{p_{h}}, \mu_{p_{h}} \rangle \\
        \textbf{M}_{c0} = \langle \mu_{p_{h}}, \mu_{P_{\delta}} \rangle \\
        \textbf{M}_{00} = \langle  \mu_{P_{\delta}}, \mu_{P_{\delta}} \rangle 
    \end{align}
\end{subequations}
From equation \ref{cone_dist} and \ref{mu_delta}, we get the following 
\begin{subequations}

    \begin{align}
        \textbf{M}_{cc} = \langle  \sum_{p=1}^{p=N^2}\textbf{a}_p k(\textbf{h}_p,.), \sum_{p=1}^{p=N^2}\textbf{a}_p k(\textbf{h}_p,.)\rangle  \\
        \textbf{M}_{c0} = \langle \sum_{p=1}^{p=N^2}\textbf{a}_p k(\textbf{h}_p,.)\sum_{q=1}^{q=N^2}\textbf{b}_q k(0,.)\rangle \\
        \textbf{M}_{00} = \langle\sum_{q=1}^{q=N^2}\textbf{b}_q k(0,.),\sum_{q=1}^{q=N^2}\textbf{b}_q k(0,.)\rangle
    \end{align}

\end{subequations}
Applying kernel trick on the above equations, we get
\begin{subequations}
    \begin{align}
        \textbf{M}_{cc} = \textbf{a}_{k}^T \textbf{K}_{cc} \textbf{a}_{k} \\
        \textbf{M}_{c0} = \textbf{a}_{k}^T \textbf{K}_{c0} \textbf{b}_{k} \\
        \textbf{M}_{00} = \textbf{b}_{k}^T \textbf{K}_{00} \textbf{b}_{k}
    \end{align}
    \begin{equation}
    \begin{split}
            l_{dist}(p_{h}, p_{\delta}) = \textbf{a}_{k}^T \textbf{K}_{cc} \textbf{a}_{k} + \textbf{a}_{k}^T \textbf{K}_{c0} \textbf{b}_{k}  + \textbf{b}_{k}^T \textbf{K}_{00} \textbf{b}_{k}^T
    \end{split}
    \end{equation}
\end{subequations}

 $\textbf{K}_{cc}$, $\textbf{K}_{c0}$ and $\textbf{K}_{00}$ are the kernel matrices and are defined as:
 
\begin{subequations}

    \begin{equation}
        \textbf{K}_{cc}=\begin{bmatrix}
            k(h_{11},h_{11}) & k(h_{11},h_{12}) & \dots & k(h_{11},h_{nn})\\
            k(h_{12},h_{11}) & k(h_{12},h_{12}) & \dots & k(h_{12},h_{nn})\\
            \vdots & \vdots & \vdots & \vdots\\
            k(h_{nn},h_{11}) & k(h_{nn},h_{12}) & \dots & k(h_{nn},h_{nn})
        \end{bmatrix}
    \end{equation}
    \begin{equation}
        \textbf{K}_{c0}=\begin{bmatrix}
            k(h_{11},0) & k(h_{11},0) & \dots & k(h_{11},0)\\
            k(h_{12},0) & k(h_{12},0) & \dots & k(h_{12},0)\\
            \vdots & \vdots & \vdots & \vdots\\
            k(h_{nn},0) & k(h_{nn},0) & \dots & k(h_{nn},0)
        \end{bmatrix}
    \end{equation}
    \begin{equation}
        \textbf{K}_{00}=1_{N^2 x N^2}
    \end{equation}
\end{subequations}
The computation time of evaluating MMD or $l_{dist}$ depends mainly on the computation time of the upper triangle of the symmetric matrix $\textbf{K}_{cc}$ as the matrix $\textbf{K}_{00}$ is a set of ones, and $\textbf{K}_{c0}$ is a column matrix.

%% file: chapters/results.tex
\section{Results and Discussion}
\label{Results}



\begin{figure*}
    \begin{subfigure} {.33\textwidth}
            \includegraphics[width=.9\textwidth]{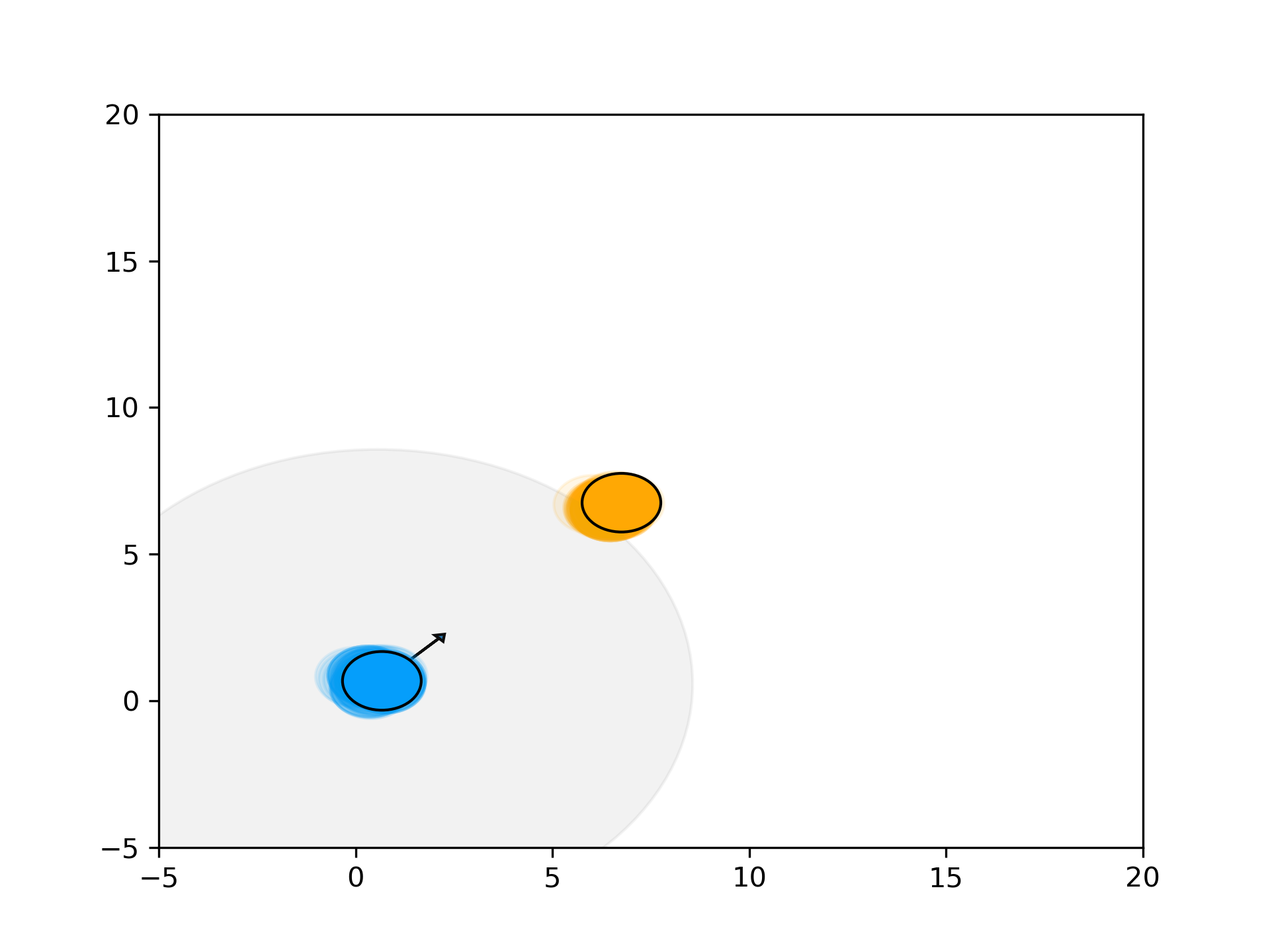}\hfill
            \caption{\small \rightskip=0.5cm Beginning of collision avoidance. The VO constraint violation distribution is very far from Dira-Delta. }
            \label{dist_match1}
    \end{subfigure}
    \begin{subfigure} {.33\textwidth}
            \includegraphics[width=.9\textwidth]{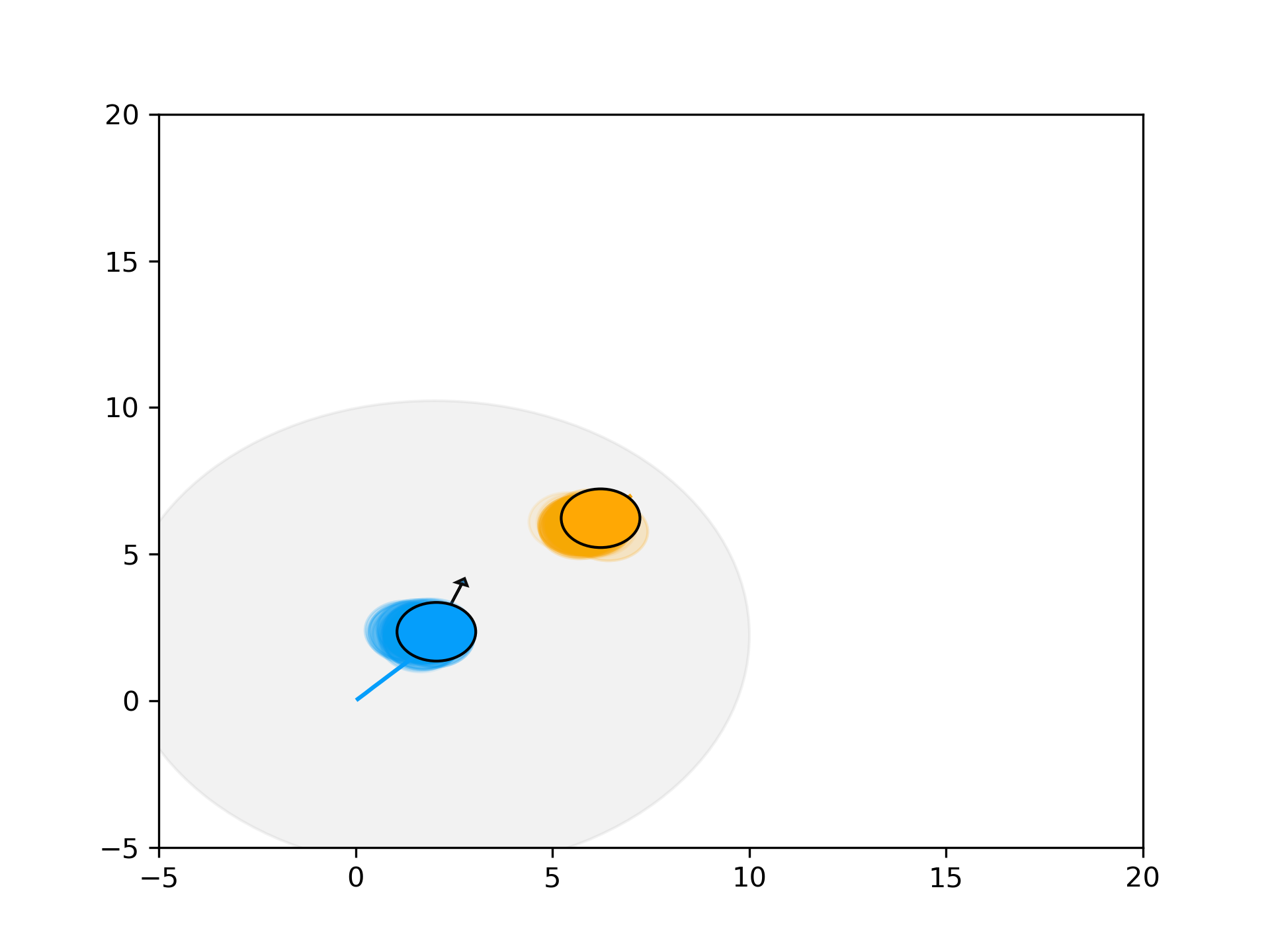}\hfill
            \caption{\small \rightskip=0.5cm As avoidance maneuver begins, some part of the VO constraint violation distribution coincides with the Dirac Delta Distribution}
            \label{dist_match2}
    \end{subfigure}
    \begin{subfigure} {.33\textwidth}
            \includegraphics[width=.9\textwidth]{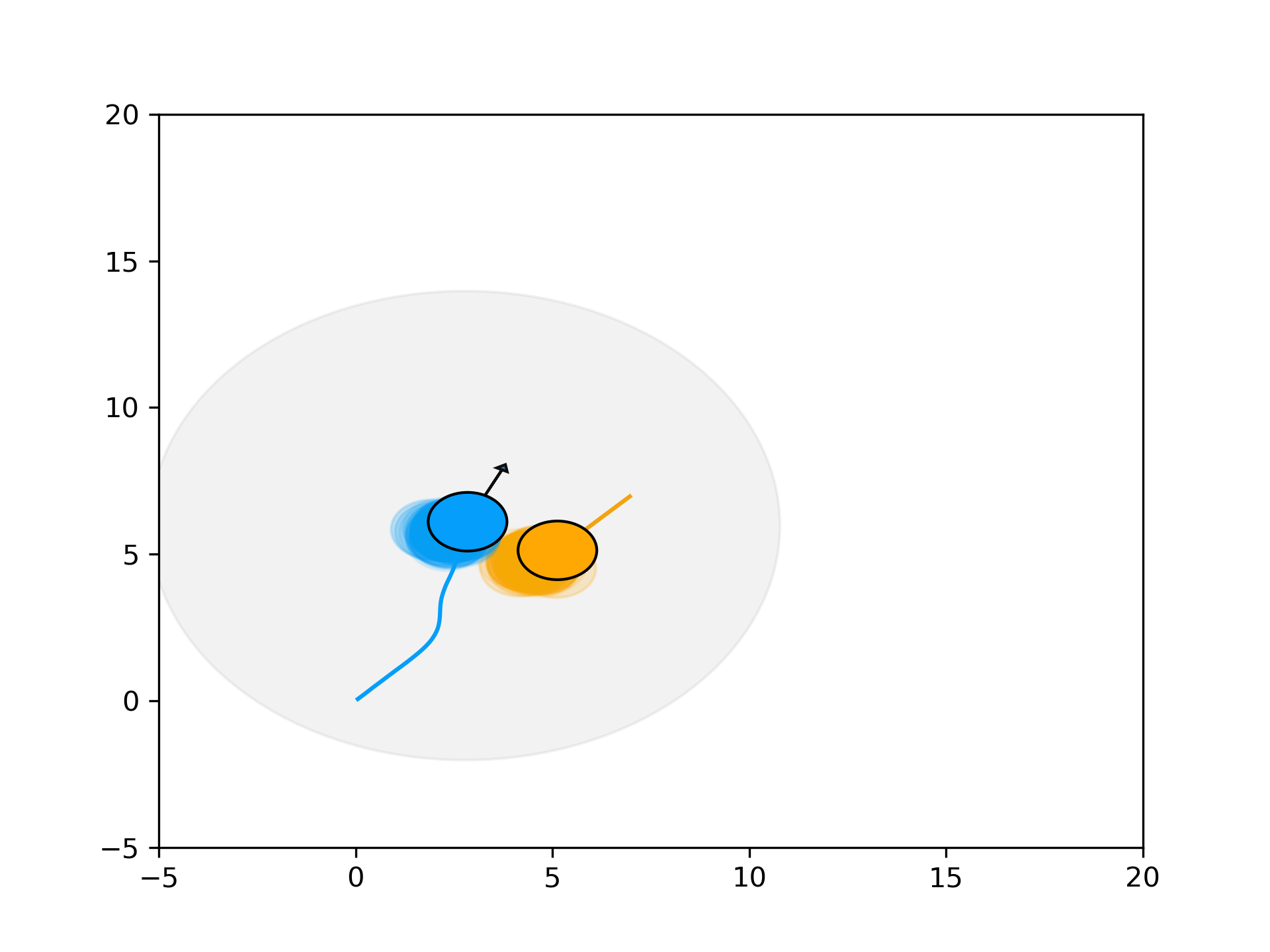}\hfill
            \caption{\small \rightskip=0.5cm Towards the end of the collision avoidance maneuver, the distribution of the VO constraint violations becomes close to that of the Dirac-Delta }
            \label{dist_match3}
    \end{subfigure}
    \begin{subfigure} {.33\textwidth}
            \includegraphics[width=.9\textwidth]{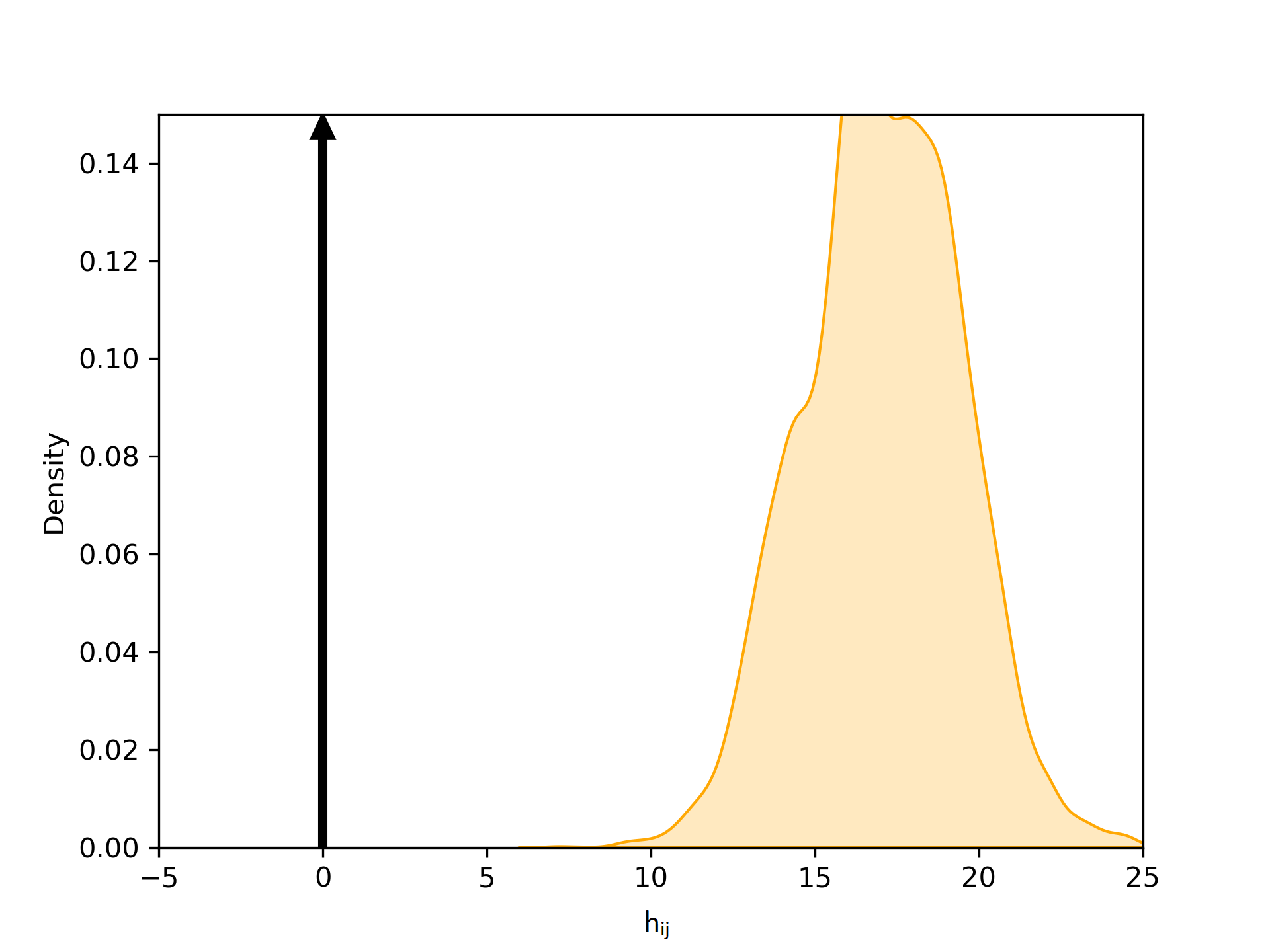}\hfill
    \end{subfigure}
    \begin{subfigure} {.33\textwidth}
            \includegraphics[width=.9\textwidth]{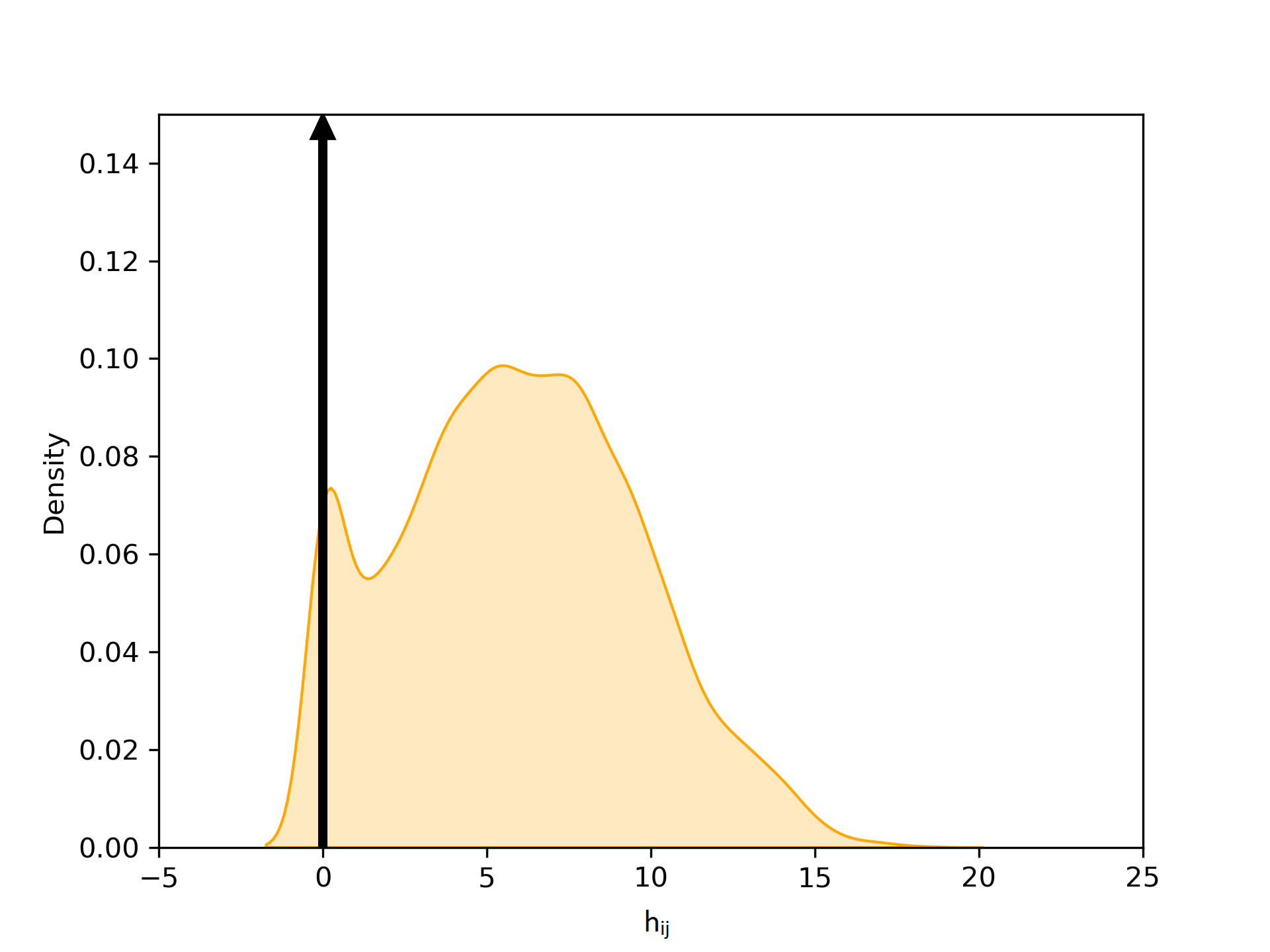}\hfill
    \end{subfigure}
    \begin{subfigure} {.33\textwidth}
            \includegraphics[width=.9\textwidth]{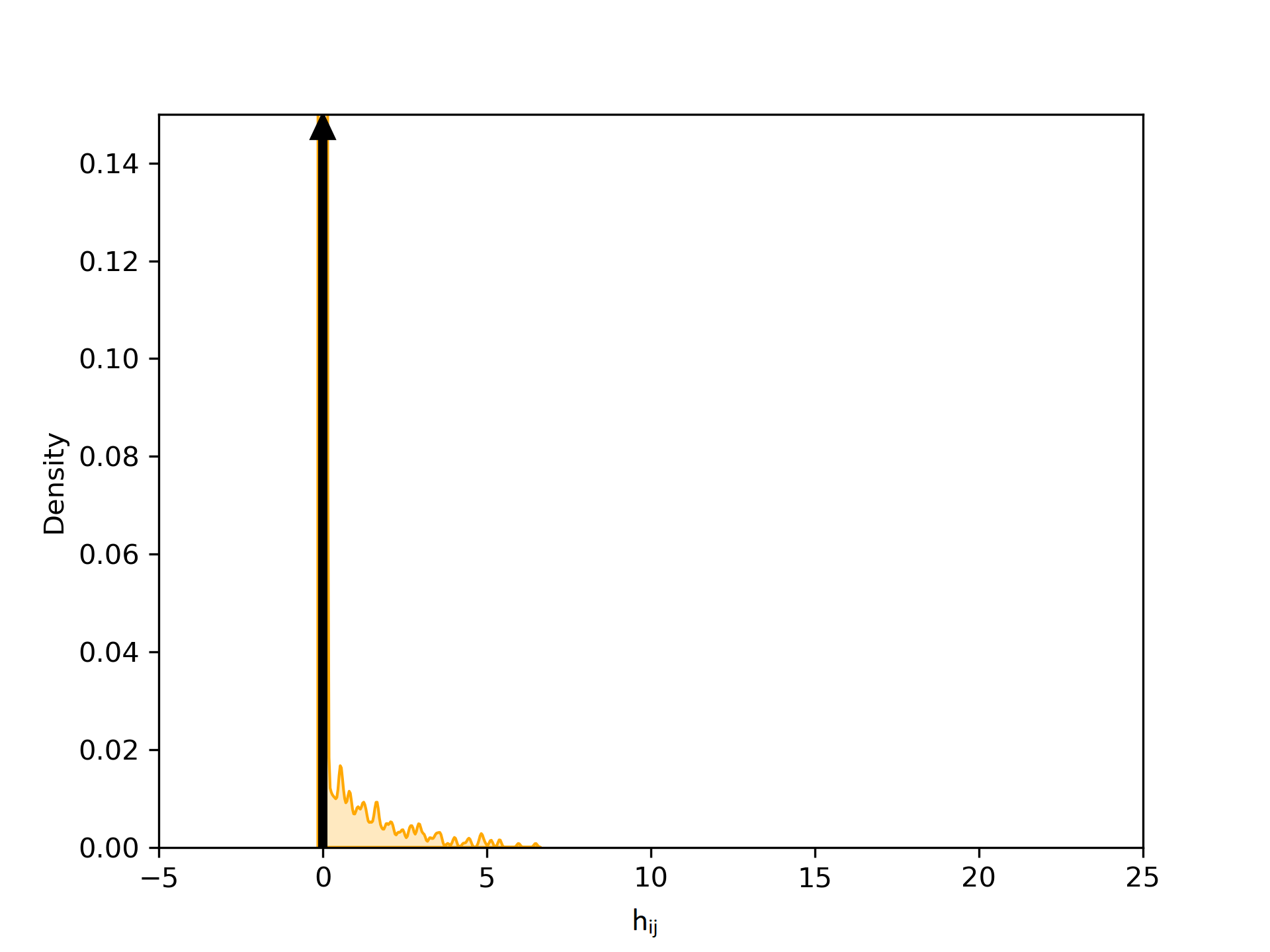}\hfill
    \end{subfigure}
    \caption{\small Validation of distribution matching interpretation of CCO. \normalsize}
    \label{dist_match}
\end{figure*}


\begin{figure*}
    \begin{subfigure}{.33\textwidth}
        \includegraphics[width=.9\textwidth]{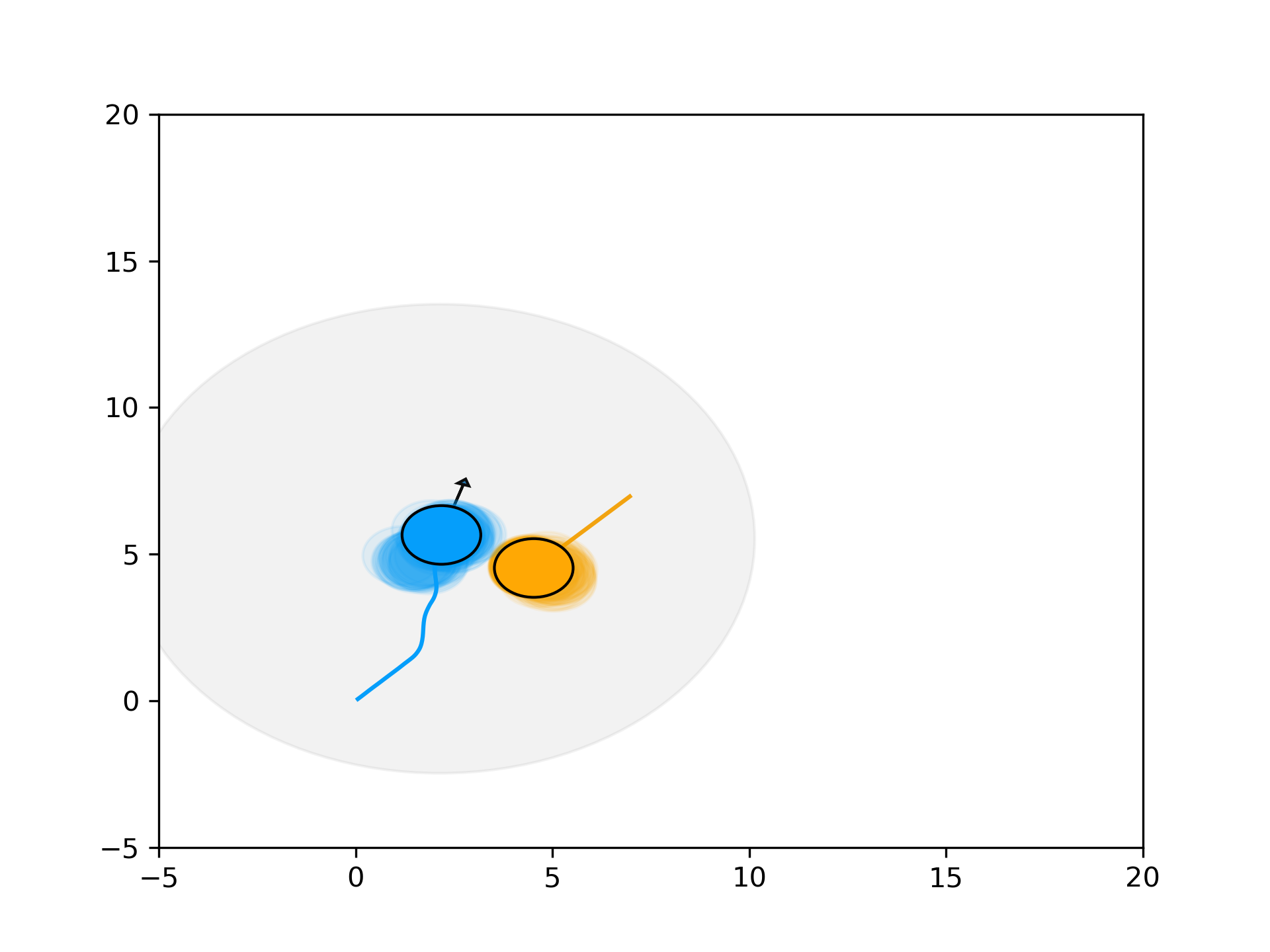}
        \caption{\small Favorable homotopy chosen by the robot}
        \label{side selection correct}
    \end{subfigure}
    \begin{subfigure}{.33\textwidth}
        \includegraphics[width=.9\textwidth]{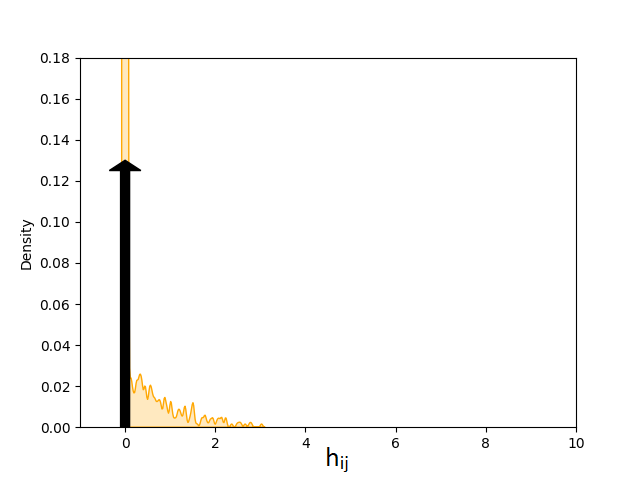}
        \caption{\small Corresponding VO constraint violation distribution}
        \label{side selection correct non gaussian}
    \end{subfigure}
    \begin{subfigure}{.33\textwidth}
        \includegraphics[width=.9\textwidth]{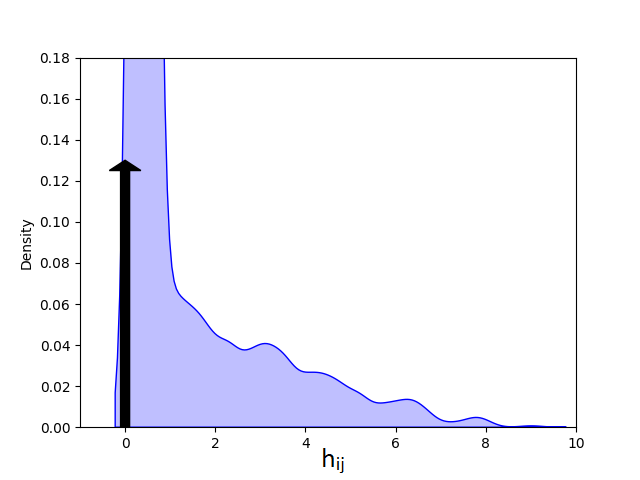}
        \caption{\small Corresponding VO constraint violation distribution under Gaussian approximation}
        \label{side selection correct gaussian}
    \end{subfigure}
    \begin{subfigure}{.33\textwidth}
        \includegraphics[width=.9\textwidth]{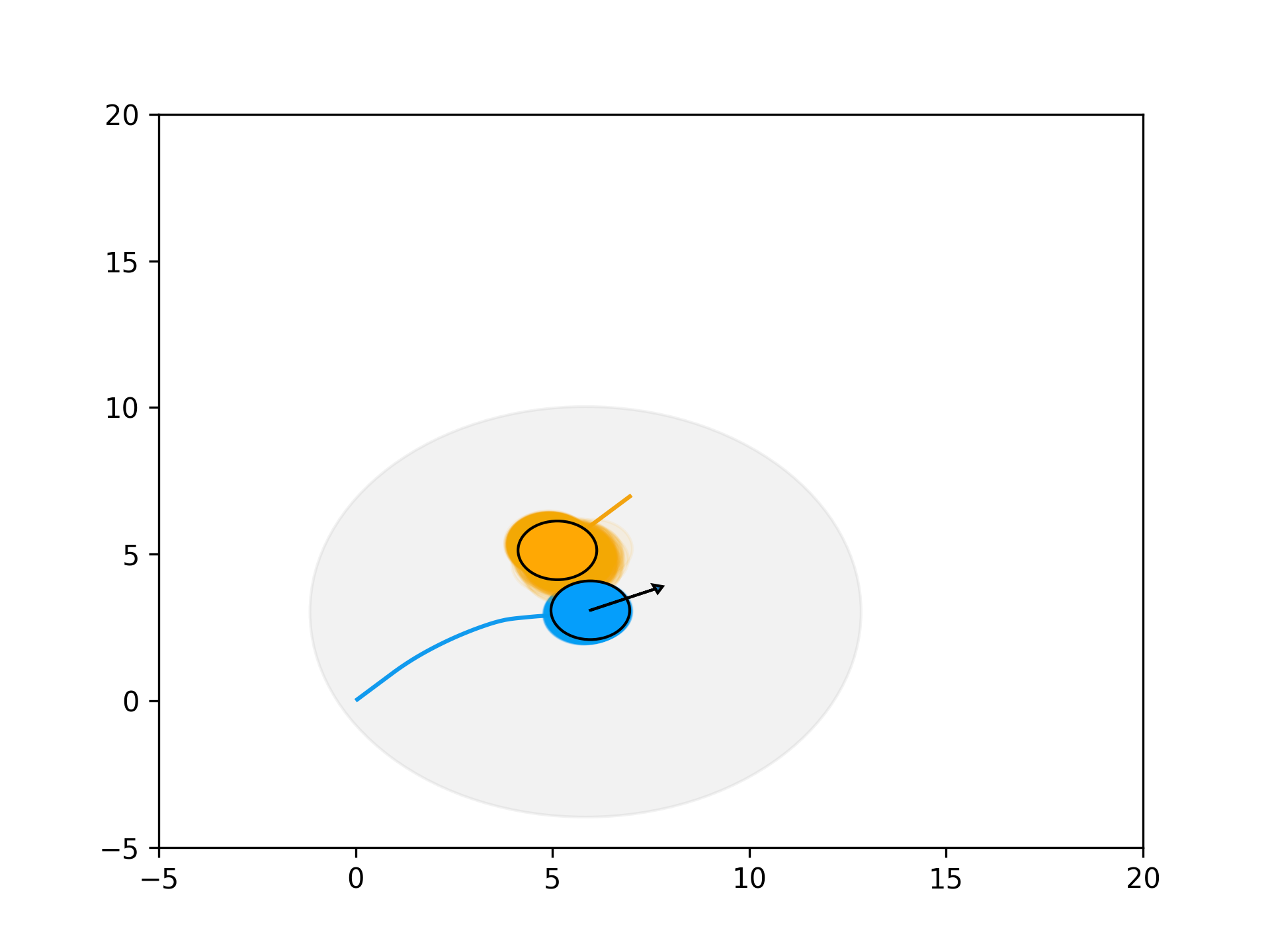}
        \caption{\small Un-favorable homotopy chosen by the robot.}
        \label{side selection incorrect}
    \end{subfigure}    
    \begin{subfigure}{.33\textwidth}
        \includegraphics[width=.9\textwidth]{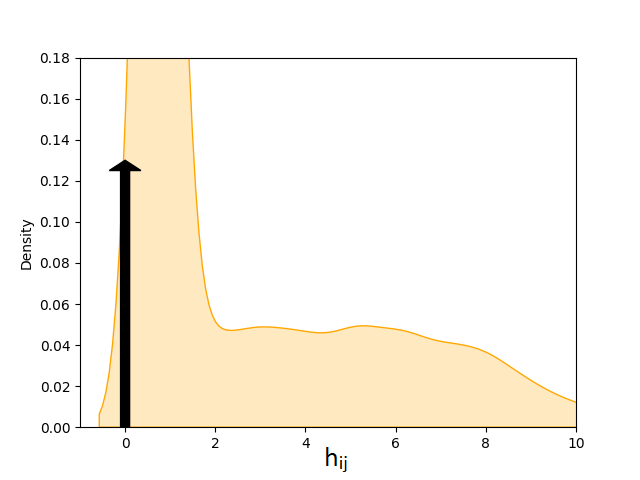}
        \caption{\small Corresponding VO constraint violation distribution}
        \label{side selection incorrect non gaussian}
    \end{subfigure} 
    \begin{subfigure}{.33\textwidth}
        \includegraphics[width=.9\textwidth]{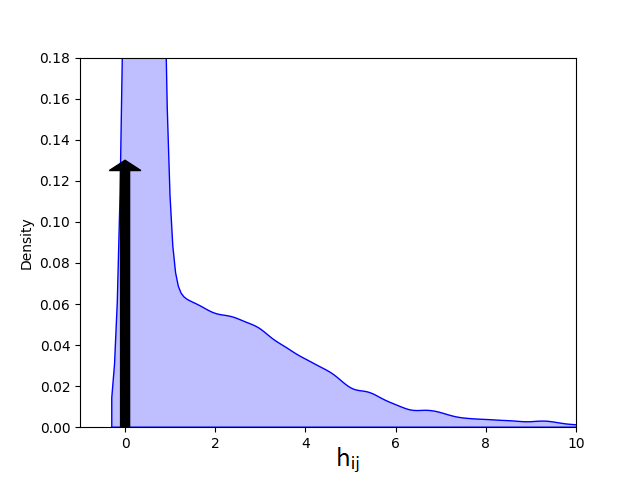}
        \caption{\small Corresponding VO constraint violation distribution under Gaussian approximation}
        \label{side selection incorrect gaussian}
    \end{subfigure}
    \caption{\small These figures illustrate how better favourable homotopy selection will lead to better distribution matching and hence, larger number of samples will be avoided  \normalsize}
     \label{side selection}
\end{figure*}

\noindent \textbf{Implementation Details:} All the simulations were carried out on a desktop in Python. The CPU and GPU used were AMD Ryzen 5 3500 and NVIDIA 1660 Super respectively. We queried 100 samples each of robot and obstacle's position and velocity from their distribution to construct the MMD $l_{dist}$ term in optimization \eqref{final_cost}-\eqref{feas_final}. We reiterate that we don't assume any knowledge on the parametric form for the underlying distribution. We used a fixed set of 625 discrete control inputs to compute the one that led to the lowest value for the cost \eqref{final_cost}. We used $\gamma=0.1$ in RBF kernel definition. In all the plots demonstrating qualitative results in the form of robot and obstacle trajectories, the blue circle represents the robot's actual position, the yellow circle represents the obstacles' position, and the lighter shade circles surrounding both of them represent the underlying uncertainty in position. Extensive qualitative results and the code can be found at \href{https://github.com/anishgupta31296/MMD-with-Dirac-Delta-Distribution}{https://github.com/anishgupta31296/MMD-with-Dirac-Delta-Distribution}.

\noindent \textbf{Baselines:} We call our approach MMD Non-Gaussian when comparing against the following baselines:

\begin{itemize}
    \item \textbf{MMD-Gaussian}: This baseline follows the same approach of distribution matching in RKHS through MMD. The only difference with our approach is that it computes a Gaussian approximation of the motion and perception noise.
    
    \item \textbf{KLD}: This baseline from \cite{bharath_ral} also follows the interpretation of CCO as a distribution matching problem. But it differs from our approach in the following respects. First, it fits a Gaussian Mixture Model to the noise distribution. Second, it works with the distribution of VO constraints, while our approach uses the distribution of violations. 
    
    \item \textbf{PVO} : This baseline from \cite{iros15_bharath} proposed a deterministic reformulation of the chance constraints over VO presented in \eqref{eq:mpc-chance-constraints}. However, it requires computing the Gaussian approximation of motion and perception noise.
\end{itemize}

\subsection{Validating Distribution Matching Interpretation}

\noindent Fig.\ref{dist_match} shows a simple scenario where a robot has an imminent head-on collision with an obstacle. Fig.\ref{dist_match1} shows at the start of the collision avoidance maneuver, the distribution of VO constraint violation is entirely on the right of zero. As the robot computes collision avoidance maneuver by solving \eqref{final_cost}-\ref{feas_final}, the constraint violations (almost) converge to the Dirac-Delta distribution. 


\begin{figure*}
     \begin{subfigure}{.24\textwidth}
        \includegraphics[width=.9\textwidth]{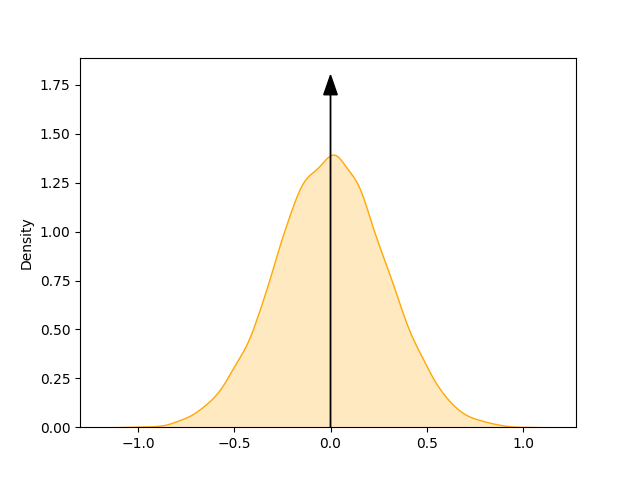}
        \caption{\small Distribution 1 }
    \end{subfigure} 
    \begin{subfigure}{.24\textwidth}
        \includegraphics[width=.9\textwidth]{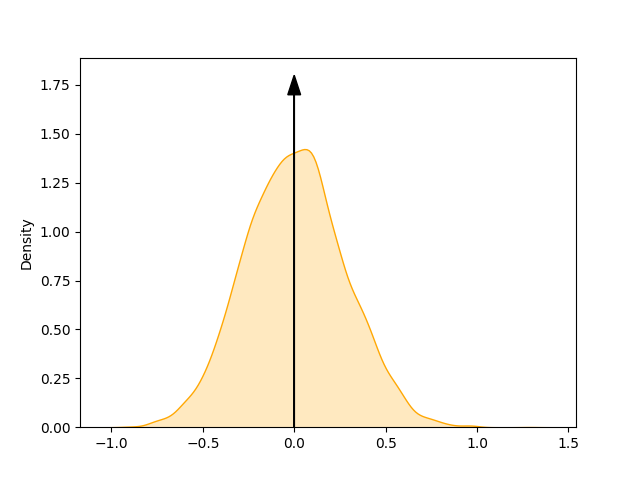}
        \caption{\small Distribution 2 }
    \end{subfigure}
    \begin{subfigure}{.24\textwidth}
        \includegraphics[width=.9\textwidth]{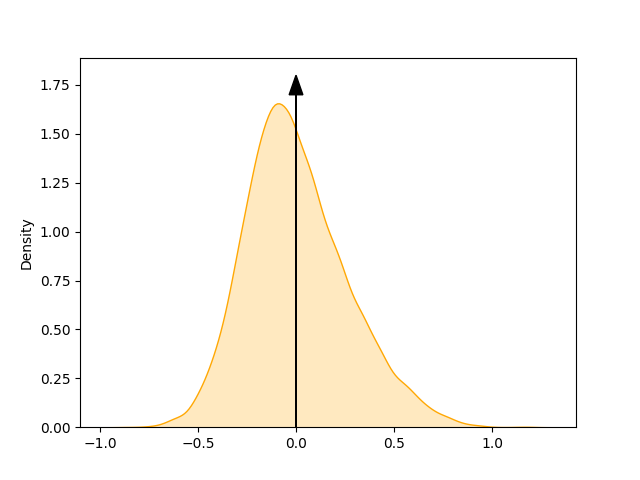}
        \caption{\small Distribution 3 }
    \end{subfigure}
    \begin{subfigure}{.24\textwidth}
        \includegraphics[width=.9\textwidth]{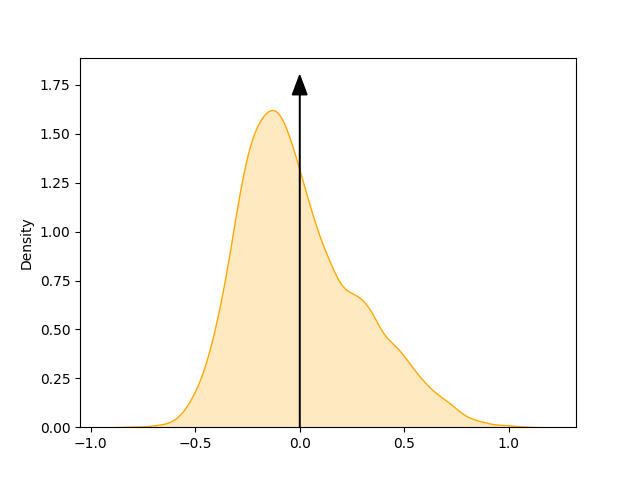}
        \caption{\small Distribution 4 }
    \end{subfigure}
    \begin{subfigure}{.24\textwidth}
        \includegraphics[width=.9\textwidth]{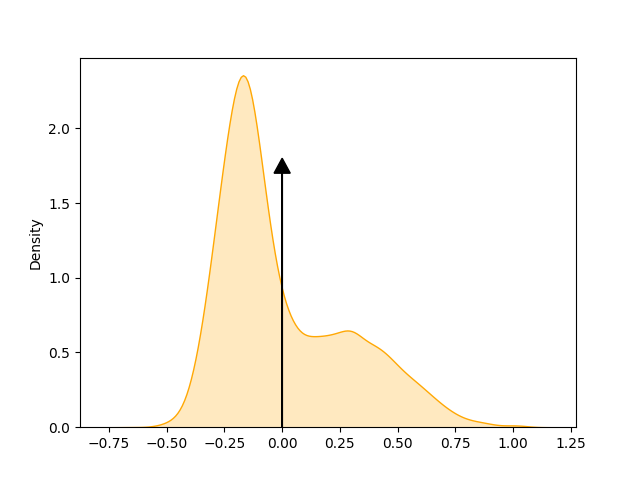}
        \caption{\small Distribution 5 }
    \end{subfigure}
    \begin{subfigure}{.24\textwidth}
        \includegraphics[width=.9\textwidth]{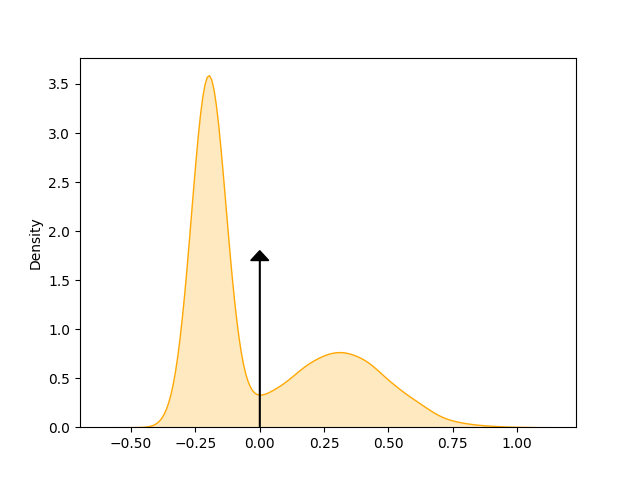}
        \caption{\small Distribution 6 }
    \end{subfigure}
     \begin{subfigure}{.24\textwidth}
        \includegraphics[width=.9\textwidth]{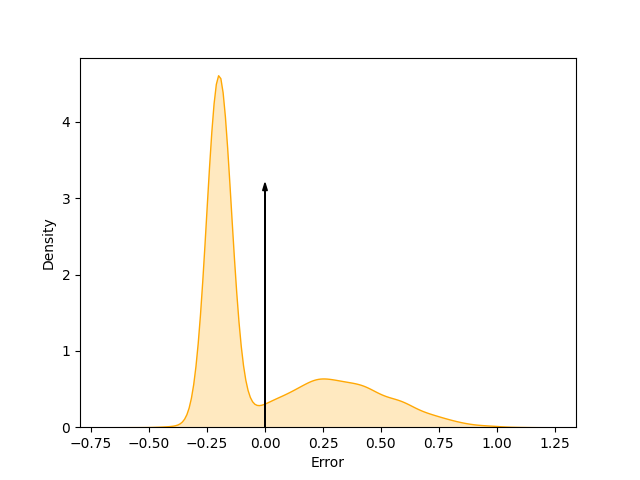}
        \caption{\small Distribution 7 }
    \end{subfigure} 
    \begin{subfigure}{.24\textwidth}
        \includegraphics[width=.9\textwidth]{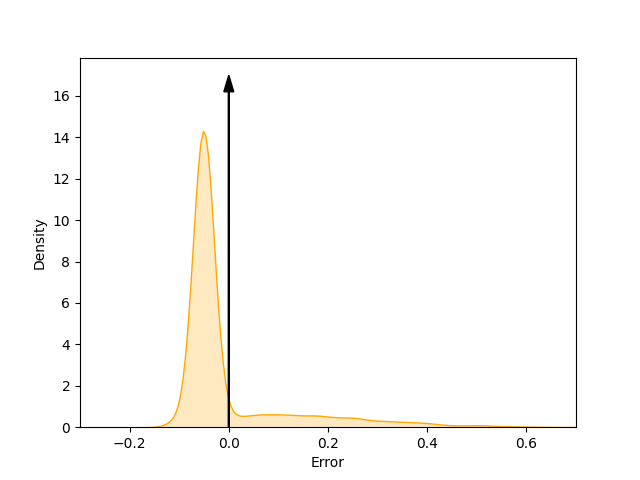}
        \caption{\small Distribution 8 }
    \end{subfigure} 
    \caption{\small Gaussian to Multimodel Transition of Distributions}
    \label{dists}
\end{figure*}

\begin{figure*}
    \begin{subfigure}{.23\textwidth}
        \includegraphics[width=.9\textwidth]{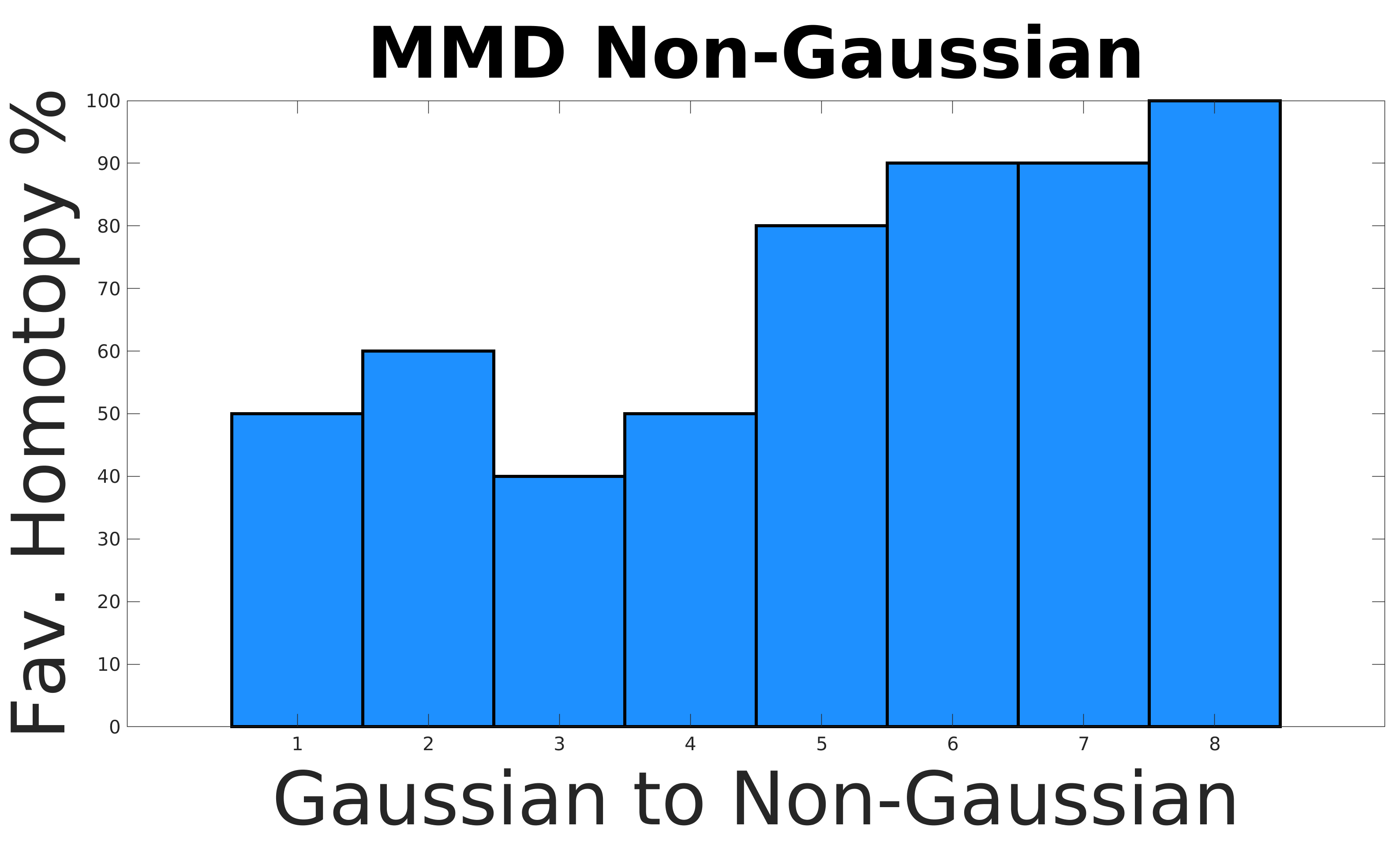}
        \caption{\small This bar plot depicts the frequency of choosing favorable homotopy with our approach MMD Non-Gaussian in a single obstacle benchmark shown in Fig.\ref{side selection correct non gaussian},\ref{side selection correct gaussian} under the noise distributions from Figure \ref{dists}. We can observe the increasing likelihood of choosing the favorable homotopy as we move towards non-Gaussian noise distributions}
        \label{non gaussian quant homotopy}
    \end{subfigure}
    \begin{subfigure}{.23\textwidth}
        \includegraphics[width=.9\textwidth]{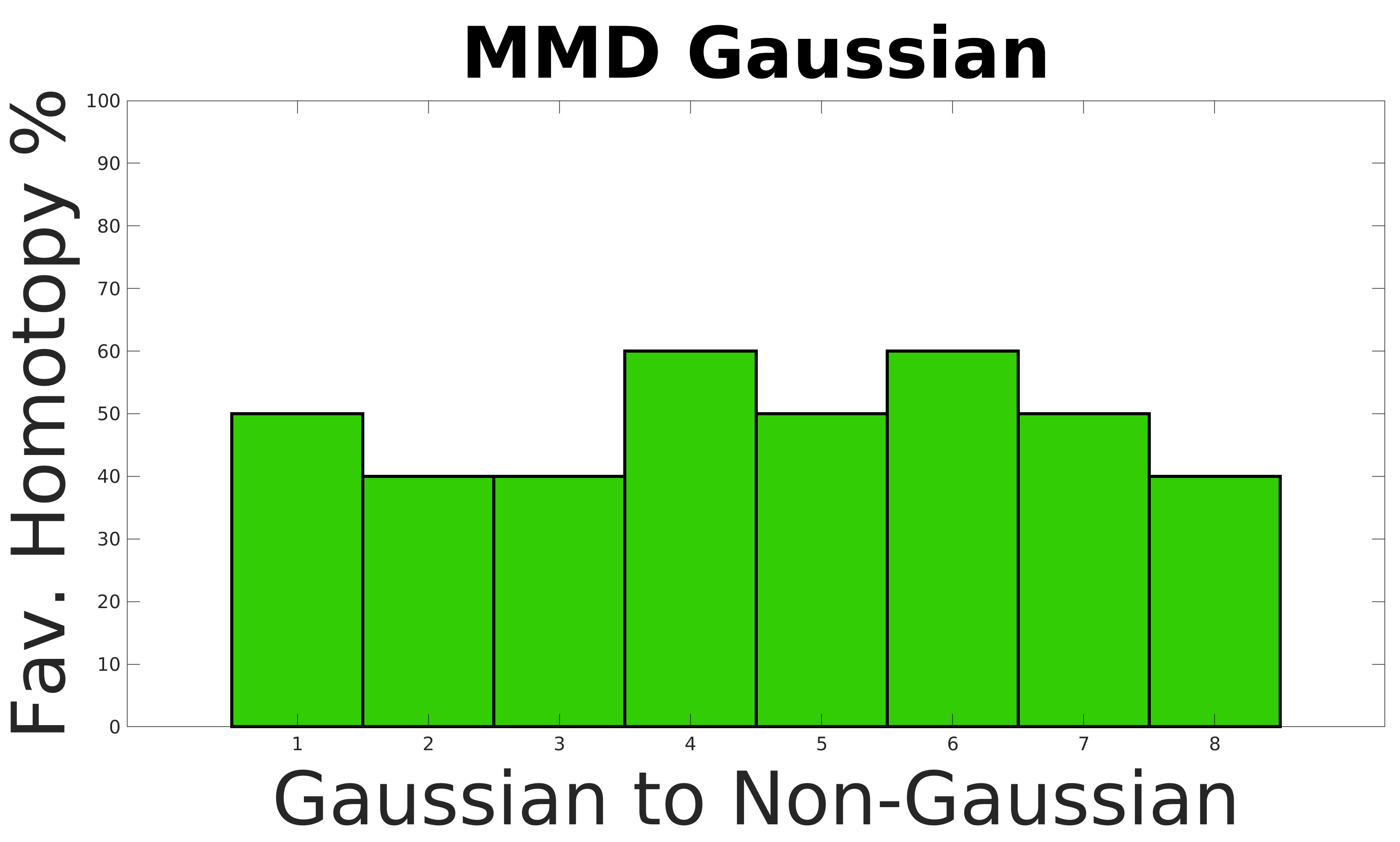}
        \caption{\small This bar plot depicts the frequency of choosing favourable homotopy for MMD Gaussian in case of single obstacle benchmark shown in Fig.\ref{side selection correct non gaussian},\ref{side selection correct gaussian} using the noise distributions in Figure \ref{dists}. We can see the probability of choosing the correct side remains similar even when the noise distribution becomes increasingly non-Gaussian.   }
        \label{gaussian quant homotopy}
    \end{subfigure}
    \begin{subfigure}{.23\textwidth}
        \includegraphics[width=.9\textwidth]{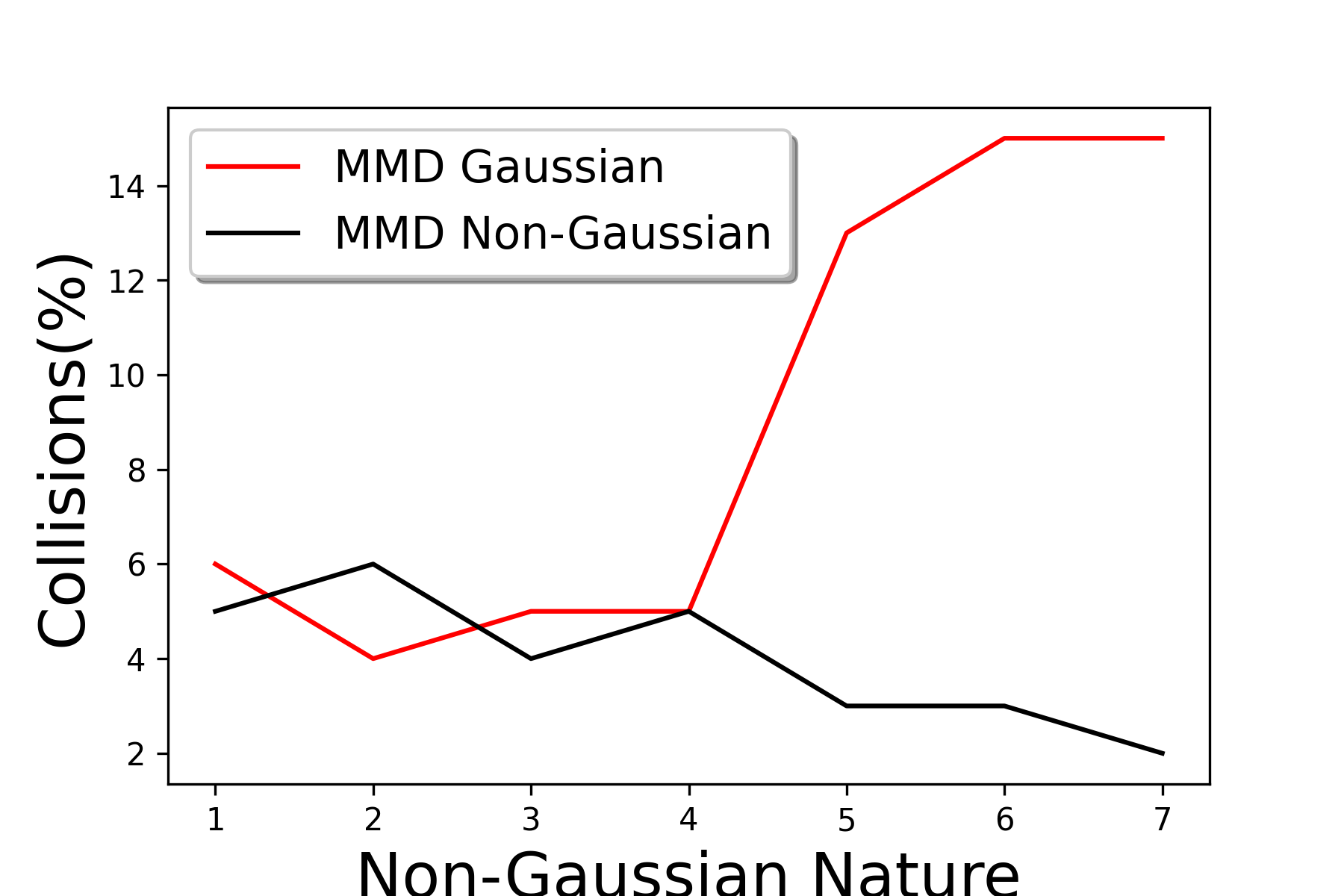}
        \caption{\small Effect of non-gaussian nature on number of samples colliding}
        \label{dist quant samples colliding}
    \end{subfigure}
    \begin{subfigure}{.23\textwidth}
        \includegraphics[width=.9\textwidth]{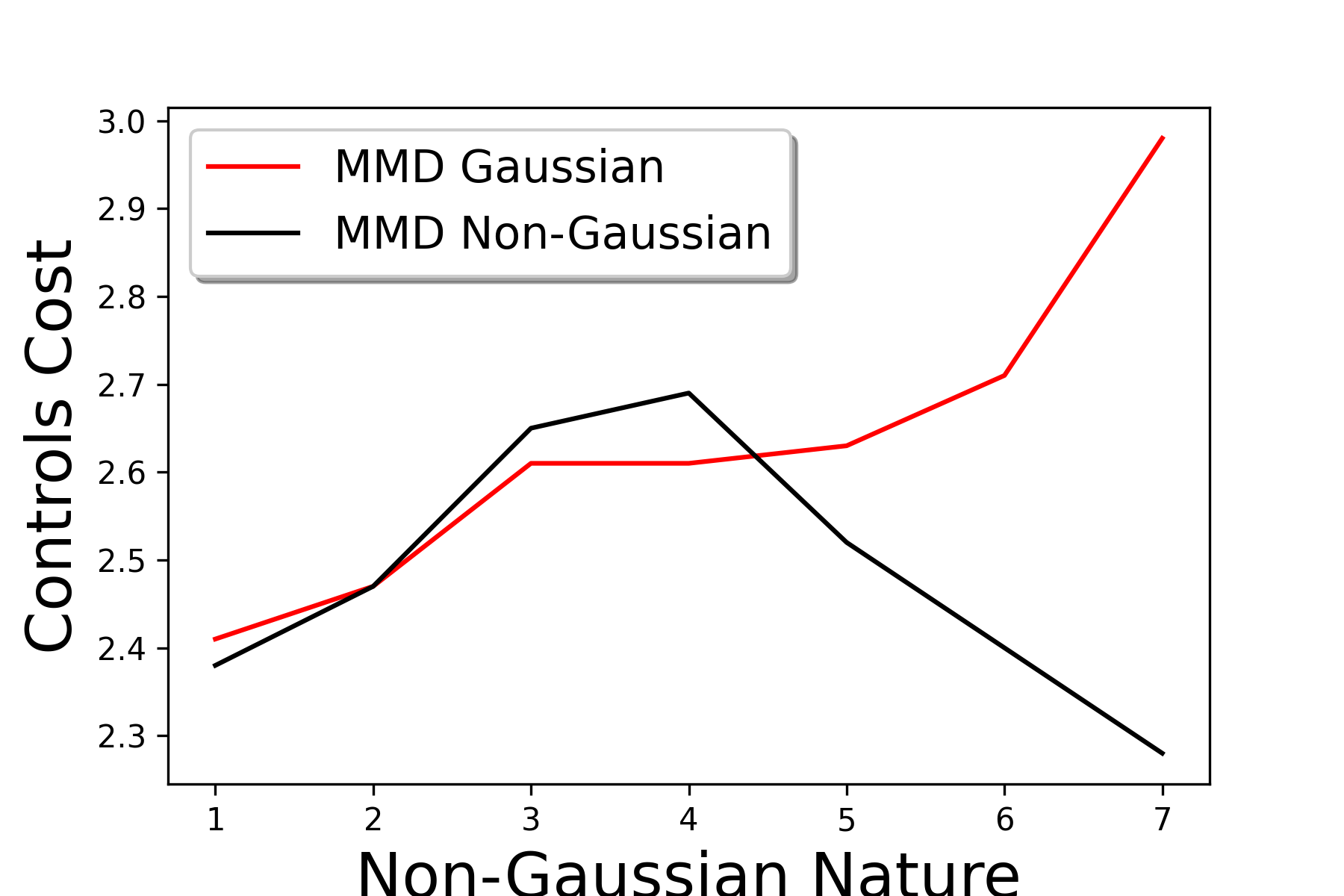}
        \caption{\small Effect of non-gaussian nature on control costs. The x-axis shows the distribution number from Fig.\ref{dists}}
        \label{dist quant control costs}
    \end{subfigure}  
    \caption{\small Quantitative Analysis on non-gaussian nature of distribution. The x-axis shows the distribution number from Fig.\ref{dists}}
    \label{dists quant}
\end{figure*}

\subsection{Analyzing the Choice of Homotopies}
\label{sec:Ablation}
\noindent This section presents the most important empirical result of our paper. We consider a benchmark with a single obstacle as shown in Fig.\ref{side selection} to analyze two key questions. First, how is the choice of homotopy related to the distribution of constraint violations for a biased non-Gaussian distribution and its Gaussian approximation. Second, we intend to study the effectiveness of our MMD Non-Gaussian approach in ensuring the selection of favorable homotopies during collision avoidance. To these ends, we sampled two control actions for the scenario shown in Fig.\ref{side selection} which results in the robot passing the obstacle from different sides. Clearly, Fig.\ref{side selection correct} is the favorable homotopy in this scenario while that shown in Fig.\ref{side selection incorrect} leads to a large overlap between the robot and obstacle position uncertainty. Fig.\ref{side selection correct non gaussian} shows the distribution of constraint violations for the control input that leads to the favorable homotopy for the true non-Gaussian distribution. It can be seen that the distribution of violation is very close to the ideal Dirac-Delta distribution. Now, contrast this with Fig.\ref{side selection incorrect non gaussian} that recreates the constraint violation distribution for the control input leading to unfavorable homotopy. We can clearly see a stark difference between Fig.\ref{side selection correct non gaussian} and \ref{side selection incorrect non gaussian}. Now, we hypothesize that any planner that can capture the true distribution of constrint violation for a given control input can easily distinguish between a favorable and unfavorable homotopy. We will soon discuss how our MMD Non-Gaussian planner in fact fits this description. But before that, we turn our attention to Fig\ref{side selection correct gaussian} and \ref{side selection incorrect gaussian} that presents the distribution of constraint violations under Gaussian approximation of the noise. As it can be seen, both favorable and un-favorable homotopy shows similar spread of the distribution mass to the right of zero. In other words, the Gaussian approximation erroneously has made both homotopies equally bad/good. As a result, it is not possible to reliably distinguish between favorable and unfavorable homotopies.

To further strengthen our claims, we design one more experiment. In Fig.\ref{dists}, we take a Gaussian distribution and then gradually make it more and more biased and multi-modal. We simulate the single obstacle avoidance benchmark of Fig.\ref{side selection} for all these noise distributions added to motion and perception. We perform 100 Monte-Carlo runs for each noise distribution using our MMD Non-Gaussian planner. Fig.\ref{non gaussian quant homotopy} shows the percentage of times the robot chooses homotopy of Fig.\ref{side selection correct} over that of Fig.\ref{side selection incorrect}. When the actual noise is Gaussian, the robot randomly chooses either homotopy. In fact for Gaussian noise, there is no real benefit provided by one homotopy over another. But as the noise becomes more and more non-Gaussian, we can clearly see a pattern emerge where the favorable homotopy is overly preferred by our planner. In contrast when we make a Gaussian approximation of the true uncertainty, this pattern is lost, as shown in Fig.\ref{gaussian quant homotopy}. Under Gaussian approximation, the robot always chooses the homotopies randomly.

Fig.\ref{dist quant samples colliding} and \ref{dist quant control costs} co-relates the right choice of homotopy to collision percentages and control cost. When the underlying noise is Gaussian, both MMD Non-Gaussian and MMD Gaussian performs similar. But as the distribution departs from Gaussian assumptions, the former outperforms the latter in both collision-rate and control costs.

\begin{table}
\caption{}
\label{tab:notations}
\begin{tabular*}{0.49\textwidth}{@{\extracolsep{\fill} }  lcc  }
\toprule
  {\textbf{Method}}  & {\textbf{Computation Time(s)}} & {\textbf{Success-Rate(\%)}} \\
\midrule
\midrule
 MMD Non-Gaussian  & 0.06  & 95.5  \\
 MMD Gaussian  & 0.07  & 72  \\
 PVO  & 0.03  & 89  \\
 KLD(GMM-fit)  & 0.06(1.44)  & 82  \\
  \hline
\end{tabular*}
\label{comp_table}
\end{table}

\subsection{Quantitative Comparisons}
\label{sec:multiobs}
\begin{figure*}
    \begin{subfigure}{.24\textwidth}
        \includegraphics[width=.9\textwidth]{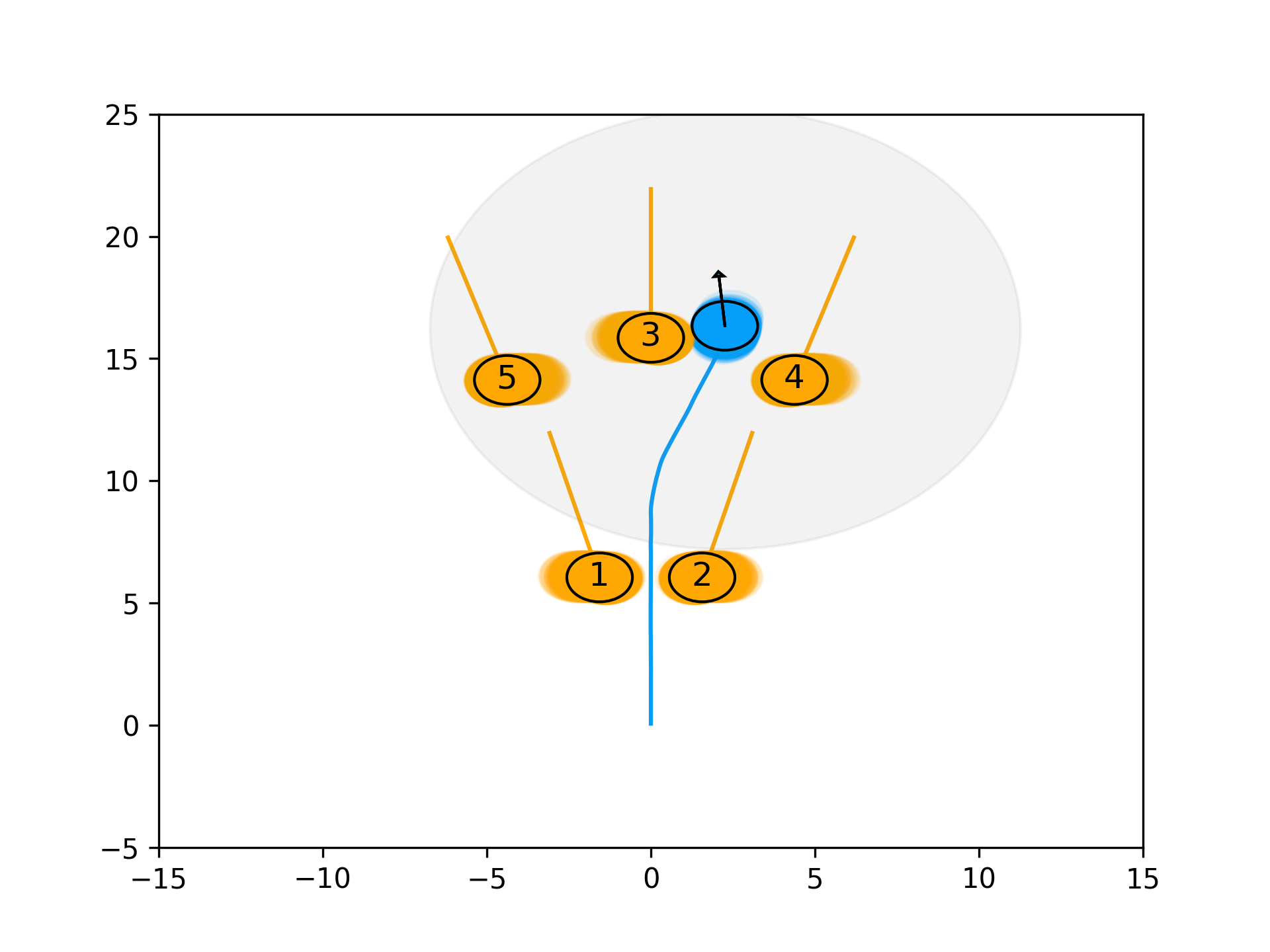}
        \caption{\small MMD Non-Gaussian}
        \label{5 obs MMD NG}
    \end{subfigure}
    \begin{subfigure}{.24\textwidth}
        \includegraphics[width=.9\textwidth]{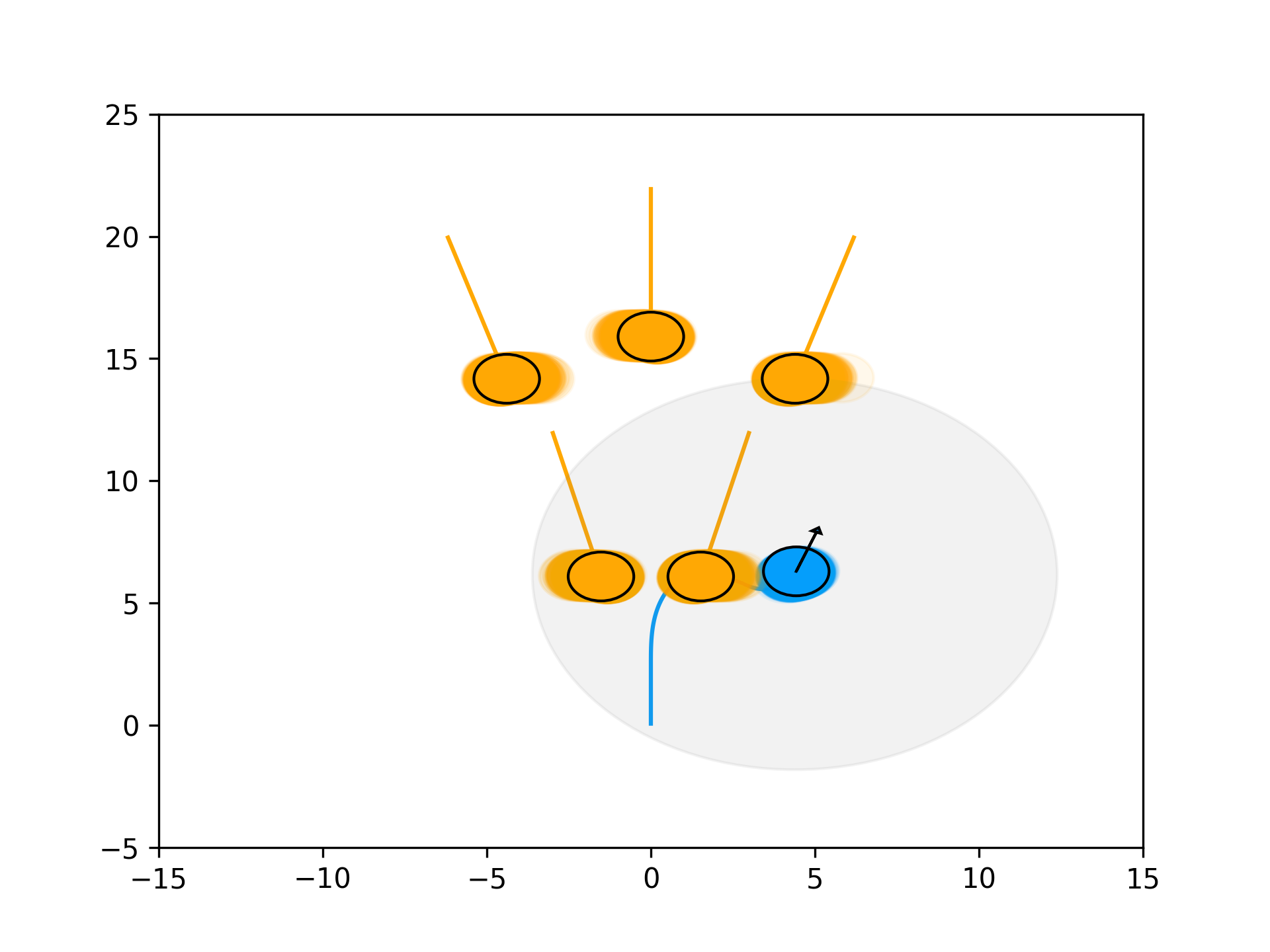}
        \caption{\small MMD Gaussian}
        \label{5 obs MMD G}
    \end{subfigure}    
    \begin{subfigure}{.24\textwidth}
        \includegraphics[width=.9\textwidth]{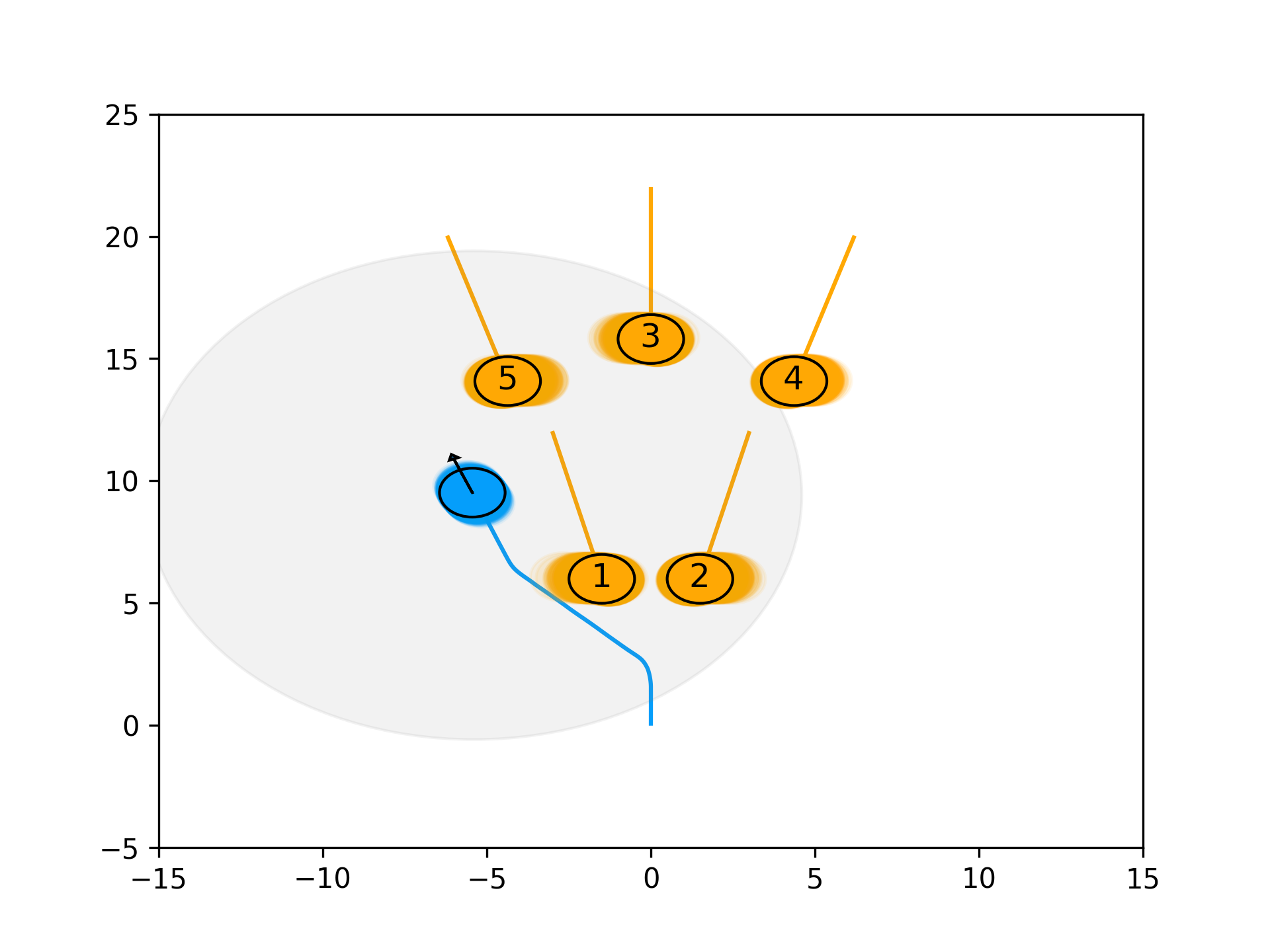}
        \caption{\small PVO}
        \label{5_obs_qualitiative}
    \end{subfigure}  
    \begin{subfigure}{.24\textwidth}
        \includegraphics[width=.9\textwidth]{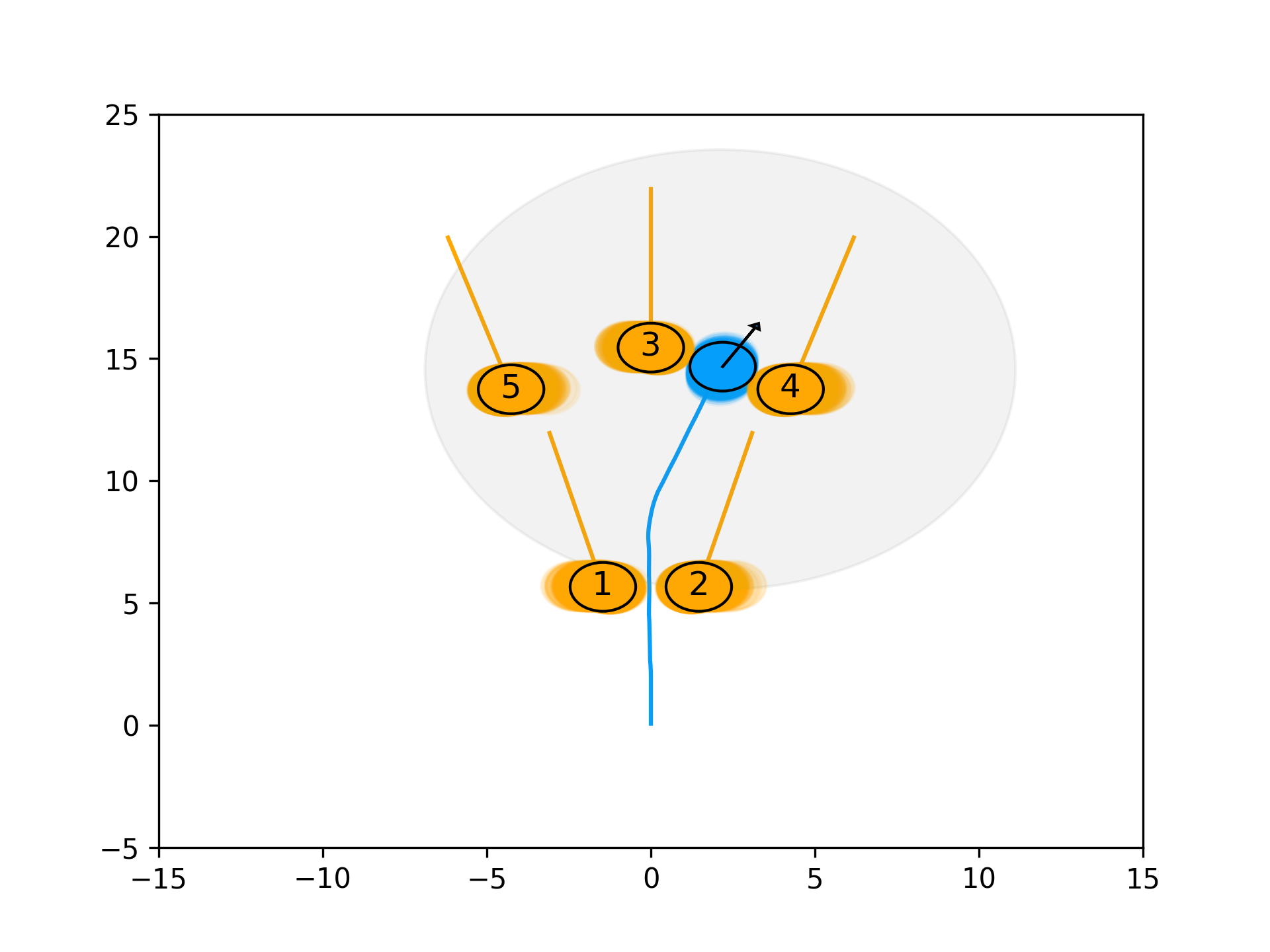}
        \caption{\small KLD}
        \label{5 obs KLD}
    \end{subfigure}  
     \caption{\small Collision avoidance using MMD Non-Gaussian and various baselines for 5 obstacle case \normalsize}
     \label{5 obs}
\end{figure*}

\begin{figure*}
    \begin{subfigure}{.33\textwidth}
        \includegraphics[width=.99\textwidth]{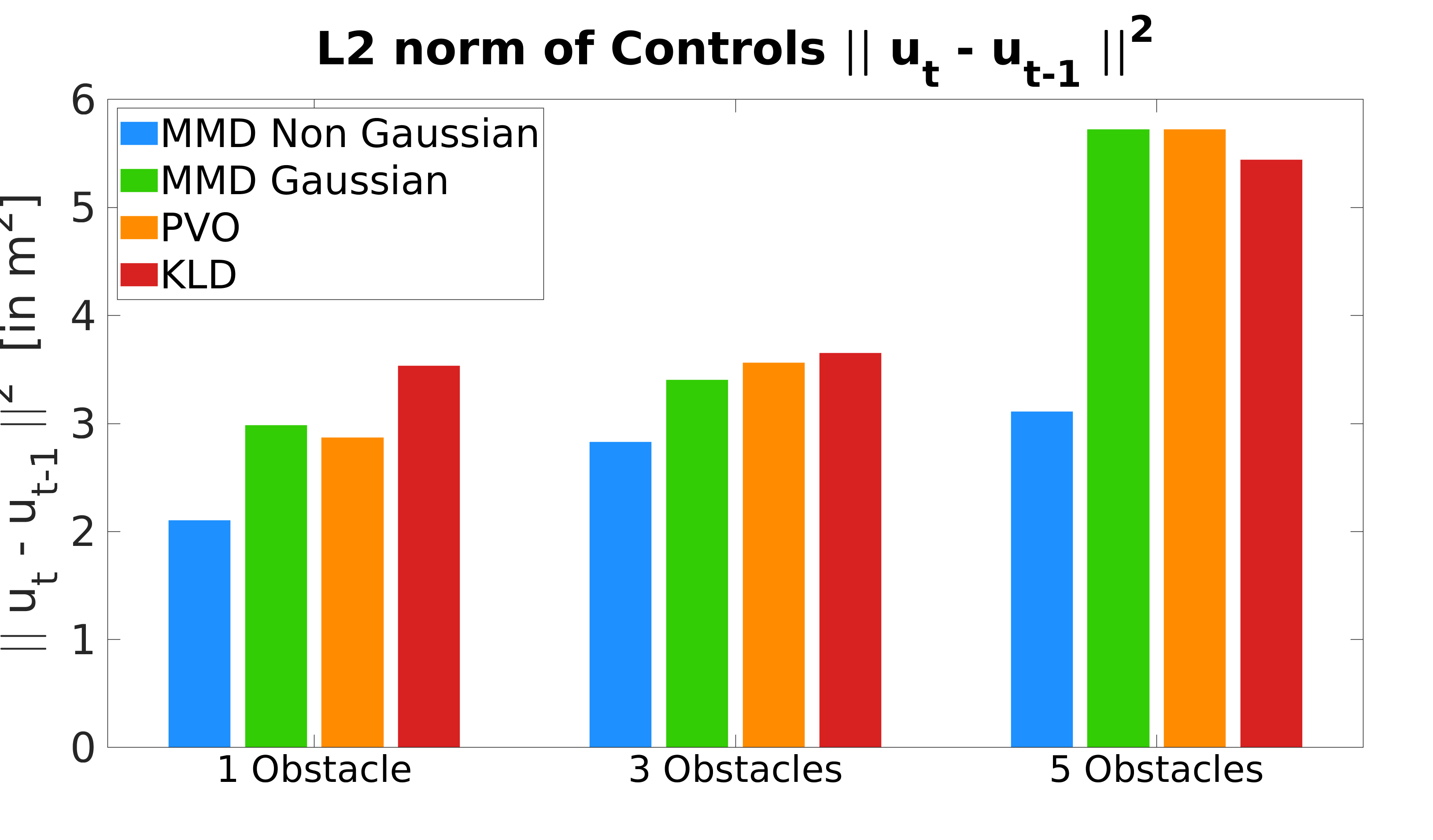}
        \caption{\small Control Costs comparison}
        \label{cc_comp}
    \end{subfigure}
    \begin{subfigure}{.33\textwidth}
        \includegraphics[width=.99\textwidth]{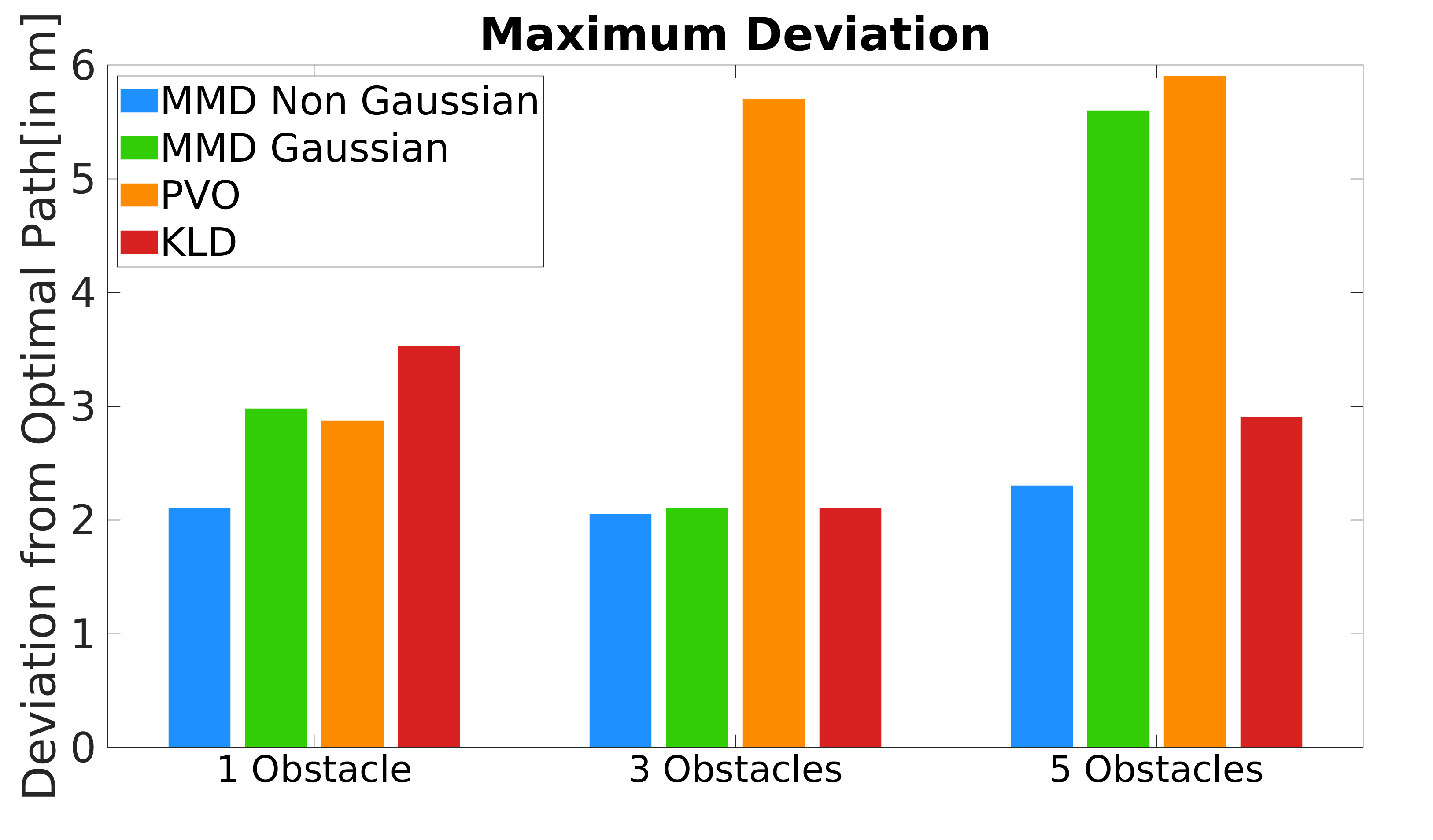}
        \caption{\small  Deviation from optimal path comparison}
        \label{dev_comp}
    \end{subfigure}    
    \begin{subfigure}{.33\textwidth}
        \includegraphics[width=.99\textwidth]{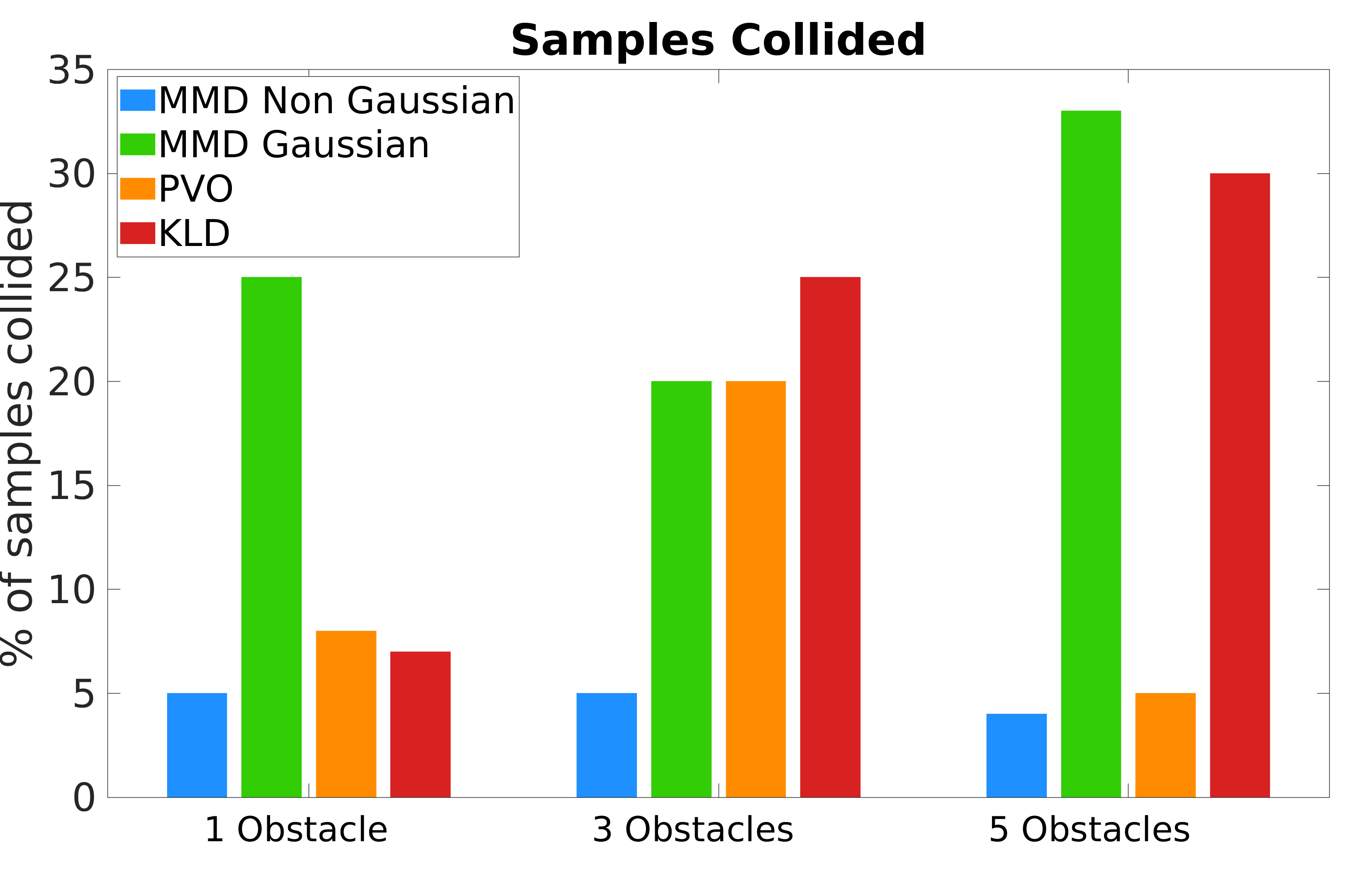}
        \caption{\small Number of colliding samples}
        \label{coll_comp}
    \end{subfigure}  
    \caption{\small  Quantitative comparison with baselines. Our approach MMD Non-Gaussian outperforms other approaches in smoothness (a), deviation from straight-line path (b) and collision probability (c) metric.}
     \label{quant_comparison}
\end{figure*}

In this section, we compare our MMD Non-Gaussian formulation with MMD Gaussian, KLD and PVO baselines defined in the beginning of section \ref{Results}. The comparisons are shown in the bar plots of Figure \ref{quant_comparison}. Fig.\ref{5 obs} presents the trajectories observed in a 5 obstacle benchmark for all the approaches. Our MMD Non-Gaussian is able to leverage the bias of the distribution and guide the robot towards homotopies that goes between the obstacles but yet has minimal overlap of robot and obstacle position uncertainty. In contrast,  both MMD Gaussian and PVO that works with Gaussian approximation of noise forces the robot to take a larger detour. This is because the Gaussian approximation over-approximates the spread of the uncertainty on either side of the robot mean position. The KLD method shows a very similar approach since it can fits a complicated a GMM to the motion and perception noise.

Figure \ref{cc_comp} compares over $L_2$ norm of control change over two consecutive instances $\Vert \textbf{u}_t-\textbf{u}_{t-1}\Vert_2^2$ which can be used to infer the smoothness of a collision avoidance maneuver. Our approach has the lowest change while all other baselines have similar trends. Fig.\ref{dev_comp} shows the comparison between the deviation that the robot exhibits from an optimal straight line path to the goal. On an average our approach is $72.84\%$ better than all the other baselines. Finally, we compare how many of the drawn position samples from the robot uncertainty collide with that of the obstacles for all the baselines. This metric serves as a proxy of collision probability. Our approach consistently maintains the percentage value at $5$ or less. All other baselines performance varies over the benchmarks and lies between $11-28\%$. This is further reiterated in Table \ref{comp_table}

Table \ref{comp_table} compares the computation time for our approach and all the baselines. The PVO approach is the fastest while the rest of the approaches have comparable run-times.

%% file: chapters/conclusion.tex
\section{Conclusion}
The paper used the distribution matching interpretation of CCO to formulate reactive dynamic obstacle avoidance as minimizing the deviation of the distribution of constraint violations from Dirac-Delta. We used this paradigm to analyze how bias in the  non-Gaussian motion and perception noise can be leveraged to choose favorable homotopies for collision avoidance. We also showed  how Gaussian approximation of the uncertainty erroneously makes all homotopies equally bad/good, thus forcing the planner to choose sub-optimal motions. We plan to extend our method to an Model Predictive Control setting in future.


